\documentclass[letterpaper]{article}
\usepackage{mypaper}
\usepackage{times}
\usepackage{helvet}
\usepackage{courier}
\usepackage[hyphens]{url}
\usepackage{graphicx}
\usepackage{natbib}
\usepackage{caption}
\usepackage{algorithm}
\usepackage{algorithmic}
\usepackage{rotating}
\usepackage{newfloat}
\usepackage{listings}
\usepackage{sectionnav}
\DeclareCaptionStyle{ruled}{labelfont=normalfont,labelsep=colon,strut=off}
\floatstyle{ruled}
\newfloat{listing}{tb}{lst}{}
\floatname{listing}{Listing}

\usepackage{xspace}
\usepackage{threeparttable}
\usepackage{hhline}
\usepackage{multirow}
\usepackage{booktabs}
\usepackage{makecell}
\usepackage{tcolorbox}
\usepackage{colortbl}
\usepackage{tabularx}
\usepackage{amsmath,amssymb,amsfonts}
\usepackage{pifont}

\newcolumntype{P}[1]{>{\centering\arraybackslash}p{#1}}

\newtheorem{definition}{Definition}
\makeatletter
\def\@opargbegintheorem#1#2#3{%
  \trivlist
  \item[\hskip \labelsep{\bfseries #1\ #2}]%
  {\normalfont\bfseries\ (#3)}\ \itshape}
\makeatother

\definecolor{mygray}{gray}{.9}
\definecolor{newref}{rgb}{0.776,0.184,0.486}
\definecolor{newblue}{rgb}{0.21,0.49,0.74}
\definecolor{boxbgblue}{rgb}{0.945, 0.99, 0.992}
\definecolor{boxborderblue}{rgb}{0.219, 0.455, 0.796}
\definecolor{keywordcl}{rgb}{0.639, 0.192, 0.165}
\hypersetup{
    colorlinks=true,
    citecolor=newblue,
    linkcolor=newblue,
    filecolor=magenta,      
    urlcolor=newref,
}

\newcommand{\method}{\texttt{SIGHT}\xspace}
\newcommand{\methodFull}{\underline{S}urprise-\underline{I}nduced \underline{G}eometric \underline{H}ierarchical \underline{T}racking\xspace}

\sectionnavtitle{Temporal Structure Matters for Efficient Test-Time Adaptation in WHAR}
\sectionnavline{%
  \sectionnavitem{nav:intro}{nav:related}{Sec 1: Intro}%
  \sectionnavsep
  \sectionnavitem{nav:related}{nav:prelim}{Sec 2: Related Work}%
  \sectionnavsep
  \sectionnavitem{nav:prelim}{nav:method}{Sec 3: Preliminaries}%
  \sectionnavsep
  \sectionnavitem{nav:method}{nav:exp}{Sec 4: Method}%
  \sectionnavsep
  \sectionnavitem{nav:exp}{nav:conclusion}{Sec 5: Experiments}%
  \sectionnavsep
  \sectionnavitem{nav:conclusion}{nav:appendix}{Sec 6: Conclusion}%
  \sectionnavsep
  \sectionnavitem{nav:appendix}{}{Appendix}%
}

\title{Temporal Structure Matters for Efficient Test-Time Adaptation\\in Wearable Human Activity Recognition}
\author{
    Zishu Zhou, Zaipeng Xie\thanks{Corresponding Author: Zaipeng Xie.}, Xuanyao Jie
}
\affiliations{
    College of Computer Science and Software Engineering, Hohai University\\
    \vspace{0.02in}
    \{zhouzishu, zaipengxie, xyjie\}@hhu.edu.cn
}

\begin{document}

\maketitle

\begin{abstract}
Wearable human activity recognition (WHAR) models often suffer from performance degradation under real-world cross-user distribution shifts. Test-time adaptation (TTA) mitigates this degradation by adapting models online using unlabeled test streams, yet existing methods largely inherit assumptions from vision tasks and underexploit the inherent inter-window temporal structure in WHAR streams. In this paper, we revisit such temporal structure as a feature-conditioned inference signal rather than merely an output-space smoothing prior. We derive the insight that temporal continuity and observation-induced feature deviations provide complementary cues for determining when to preserve or release temporal inertia and where to route prediction refinement during likely transitions.
Building upon this insight, we propose \textbf{\method}, a lightweight and backpropagation-free TTA framework for WHAR, enabling real-time edge deployment. \method estimates predictive surprise by comparing the current feature with a prototype-based expected state, and then uses the resulting feature deviation to guide geometry-aware transition routing based on prototype alignment and stream-level marginal habit tracking. Evaluations on real-world datasets confirm that \method outperforms existing TTA baselines while reducing computational and memory costs.
\end{abstract}

\begin{links}
    \link{Code}{https://github.com/zszhou21/SIGHT}
\end{links}

\section{Introduction}
\sectionnavtarget{nav:intro}
Wearable human activity recognition (WHAR), which infers activities from wearable sensor signals, is a fundamental task in ubiquitous computing, with applications in health monitoring, sports tracking, and human-computer interaction~\citep{WHARSurvey_IEEECST2013,CrossDomain-HAR_TIST2025}. Driven by abundant sensor data, deep learning models have achieved remarkable performance on WHAR tasks~\citep{DLWHARSurvey_CSUR2021,KD-for-WHAR_TNNLS2025}. However, these models often degrade in real-world deployment due to severe distribution shifts between training and deployment environments~\citep{DA-for-WHAR_CIKM2025}. While unsupervised domain adaptation (UDA) mitigates such shifts by exploiting labeled source and unlabeled target data~\citep{SWL-Adapt_AAAI2023}, and source-free domain adaptation (SFDA) removes source data dependence to protect privacy~\citep{MobHAR_Ubicomp2025,SF-Adapter_Ubicomp2024}, their need for concurrent source data access or a computationally intensive offline adaptation phase makes them impractical for real-time edge applications.

\begin{figure}[t]
\centering
\includegraphics[width=\columnwidth]{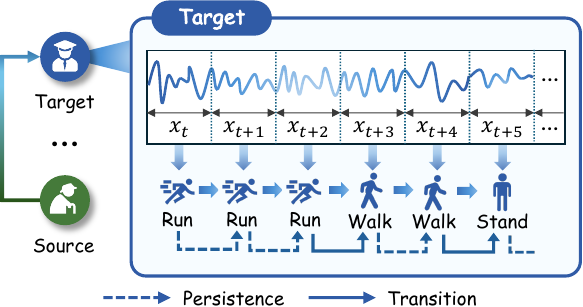}
\caption{\textbf{Scenario Illustration.} Sensor streams naturally form temporal structure through sustained activity segments and infrequent transitions between them over time.}
\label{fig:intro}
\end{figure}

Test-time adaptation (TTA) has gained attention as a promising approach to address distribution shifts by adapting models during inference using only test data~\citep{Tent_ICLR2021}. Compared to UDA and SFDA, TTA is more practical for real-world applications as it requires neither source data nor a separate adaptation phase, and is particularly appealing for WHAR where cross-subject shifts are common and devices often have limited computational resources~\citep{RealTimeHAR_BIOSTEC2023}. Recently, TTA for vision tasks has been extensively studied~\citep{REM_ICML2025,DualDomainAttribute_TIP2026,Panda_AAAI2026}, but WHAR-related TTA methods remain in their infancy, with few recent works~\citep{OFTTA_Ubicomp2024,ACCUP_KDD2025,COA-HAR_ESWA2026}. Despite these efforts, their effectiveness in practical WHAR applications remains limited. This is largely because they are built on traditional assumptions of vision TTA without fully considering the unique characteristics of WHAR scenarios. As shown in Figure~\ref{fig:intro}, sensor streams often exhibit a \textit{temporal structure}, where activities persist over consecutive windows and transitions between activities are infrequent.

In this work, we aim to develop an efficient TTA framework that fully leverages this temporal structure of WHAR streams to improve adaptation performance and efficiency. Activity persistence, i.e., adjacent windows often share the same latent activity, offers a natural prior for stabilizing noisy predictions~\citep{Markov-HAR_JMS2015,OATTA_arXiv2026}. However, real-world WHAR streams are inherently non-stationary, as users may switch activities at irregular moments~\citep{CAPTURE-24_SciData2024}, and the source model may already be biased under cross-subject shifts~\citep{CDHAR_IJBHI2025}.
Blindly enforcing temporal continuity can therefore over-preserve historical states and prevent the model from promptly adapting to emerging activities.
An effective WHAR-oriented TTA method should preserve temporal inertia in stable segments while releasing it when the current observation becomes inconsistent with the expected activity state.
This leads to our first question: \ding{182}~\textit{How can we identify when an incoming window violates temporal persistence and should trigger inertia release?}

Even when temporal inertia should be released, another challenge remains: determining where the prediction should move.
Existing temporal dynamics-aware methods~\citep{OATTA_arXiv2026} estimate class transitions in the output space using learned Markovian priors, but these statistics can be biased by noisy predictions and skewed activity frequencies.
Such an output-space prior can indicate which labels have frequently followed each other, but it cannot verify whether the current observation actually supports the predicted transition.
Moreover, output-space transition statistics ignore the local geometric evidence carried by the current sensor observation.
Since activity changes in WHAR are induced by continuous body motions~\citep{HAR-Segmentation_IEEESensors2023}, the deviation from the expected state can indicate where the prediction should move. Accordingly, our second question is: \ding{183}~\textit{How can the current observation guide online prediction refinement when a new activity transition emerges?}

In light of these two questions, we revisit temporal structure in WHAR streams as a feature-conditioned inference signal for online TTA.
To answer \ding{182}, we construct a prototype-based expected state from the previous refined prediction and compare it with the current observation, yielding a predictive surprise score that preserves temporal inertia in stable segments and releases it near likely transitions.
In response to \ding{183}, we exploit the direction of feature deviation to guide prediction refinement toward geometrically aligned activity prototypes, and maintain a habit vector to incorporate stream-level marginal activity preferences.
Integrating these mechanisms, we propose \textit{\methodFull} (\textbf{\method}), an efficient temporal structure-aware TTA framework for WHAR that enables feature-conditioned prediction refinement while remaining lightweight and backpropagation-free for edge deployment. Overall, our contributions are as follows:
\begin{enumerate}
    \renewcommand{\labelenumi}{\arabic{enumi})}
    \item We reveal that activity persistence and transitions in WHAR streams provide a feature-conditioned test-time adaptation signal for deciding both when to release temporal inertia and where to route prediction refinement under evolving target-stream dynamics.
    \item We propose \method, a lightweight, backpropagation-free TTA framework where predictive surprise controls temporal inertia, geometric attention guides transition routing, and marginal habit tracking preserves stream-level activity context for online prediction refinement.
    \item Extensive experiments on real-world and free-living WHAR datasets demonstrate that our method not only achieves superior overall performance compared to baselines, but also significantly reduces computational costs.
\end{enumerate}

\section{Related Work}
\sectionnavtarget{nav:related}

\subsection{Human Activity Recognition under Shifts}
A central challenge in wearable human activity recognition (WHAR) is the performance degradation caused by distribution shifts between the source and target domains. The inherent heterogeneity among individuals, which arises from diverse motion kinematics, unique physiological profiles, and idiosyncratic device-wearing behaviors, inevitably induces severe cross-subject distribution shifts~\citep{StudyUDAHAR_IMWUT2020,CASWL-Adapt_ECAI2024}. To address this issue, unsupervised domain adaptation (UDA) has been widely studied in WHAR to transfer knowledge from a labeled source domain to an unlabeled target domain~\citep{ContrasGAN_PMC2021}. Source-free domain adaptation (SFDA) has also been explored to further enable adaptation without access to source data to ensure user privacy and reduce storage needs. Recent SFDA methods for WHAR achieve lightweight adaptation via adapters and sample selection~\citep{SF-Adapter_Ubicomp2024}, and enable robust transfer through source--target similarity and augmentation consistency~\citep{MobHAR_Ubicomp2025}.

However, UDA requires source data, while both UDA and SFDA rely on a separate adaptation phase, imposing heavy computational burdens on resource-constrained edge devices where WHAR models are typically deployed. Although lightweight SFDA methods have been proposed, they still require multiple passes over the test set and are not tailored to real-world streaming data~\citep{OFTTA_Ubicomp2024}.

\subsection{Test-Time Adaptation}
Test-time adaptation (TTA) focuses on adapting models during inference using only incoming test data, without requiring offline adaptation or access to source data~\citep{TTA-Survey_IJCV2025}. One line of work mitigates test-time distribution shift by updating normalization statistics or normalization parameters~\citep{Tent_ICLR2021,TTN_ICLR2023}. Another dominant line performs lightweight unsupervised optimization on test data through entropy minimization or pseudo-label-based self-training~\citep{EATA_ICML2022,SAR_ICLR2023}. In addition, some methods improve adaptation stability via augmentation consistency or teacher--student frameworks~\citep{CoTTA_CVPR2022}. Optimization-free approaches also adjust predictions directly in the output space~\citep{LAME_CVPR2022,ZOTTA_arXiv2026,OATTA_arXiv2026}. Recently, TTA has been extended to WHAR through normalization-based adaptation~\citep{OFTTA_Ubicomp2024}, contrastive learning~\citep{COA-HAR_ESWA2026}, and prototype refinement~\citep{ACCUP_KDD2025}.

However, these approaches often overlook the distinctive characteristics of WHAR. Continual TTA methods mainly treat temporal correlations as a non-i.i.d.\ challenge rather than an adaptation signal~\citep{NOTE_NeurIPS2022,RoTTA_CVPR2023,MoASE_AAAI2026}, and existing WHAR-related TTA methods still largely follow the vision paradigm~\citep{ACCUP_KDD2025,COA-HAR_ESWA2026}, treating consecutive windows instance-wise and often requiring backpropagation or multiple test-time passes. Recent temporal dynamics-aware OATTA~\citep{OATTA_arXiv2026} is a step forward, but its output-space temporal statistics can be biased by noisy predictions and miss local observation-level geometric cues.

\section{Preliminaries}
\sectionnavtarget{nav:prelim}

\subsection{Problem Setup}
We consider TTA in WHAR under real-world scenarios. Let $f_\theta=h_\psi \circ g_\phi$ denote a source-trained classification model, where $g_\phi$ and $h_\psi$ are the feature encoder and linear classifier, respectively. Given an unlabeled target stream $\{x_t\}_{t=1}^{N}$, where each $x_t \in \mathbb{R}^{D \times L}$ is a multivariate sensor window with $D$ channels and $L$ time steps, the classifier first extracts a representation $z_t=g_\phi(x_t)$ and then produces the raw class distribution $p_t=\mathrm{Softmax}(h_\psi(z_t)) \in \Delta^{K-1}$. The goal is to refine this raw prediction into a more reliable online class distribution $q_t \in \Delta^{K-1}$ at each step, without access to ground-truth labels or future context. In real-world WHAR settings, the target inputs arrive sequentially and exhibit a latent temporal structure, formally defined as follows:

\begin{definition}[Temporal Structure]\label{def:temporal_structure}
A WHAR stream with latent activity state $s_t$ and feature $z_t$ exhibits inter-window temporal structure when:
(i) Activity Persistence: $s_t=s_{t-1}$ holds for most adjacent windows, yielding contiguous activity segments;
(ii) Transition Geometry: features vary mildly within the same segment but show larger, directionally informative deviations near genuine activity transitions.
\end{definition}

Notably, this inter-window temporal structure differs from intra-window temporal dependency, which is captured by the time-series backbone through sensor dynamics modeling within each window~\citep{WHAREncoder_EAAI2024,CNNHAR_IJBHI2025}.

\subsection{Classifier Weights as Initial Prototypes}
In TTA, the source data are unavailable. The pretrained linear classifier $h_\psi$ provides a compact source-side prior for class-level geometry. Let the linear head be parameterized by class weights $W=[w_1,\ldots,w_K]^\top$, where the logit of class $k$ is $\ell_{t,k}=w_k^\top z_t+b_k$. Following prior studies that use classifier weights as class anchors or to construct class prototypes~\citep{FewShotWithoutForgetting_CVPR2018,SHOT_ICML2020,T3A_NeurIPS2021}, we define the initial prototype of activity $k$ as the normalized classifier weight:
\begin{equation}
    \mu_k^{(0)} = \mathrm{Norm}{(w_k)}, \quad k=1,\ldots,K,
\end{equation}
where $\mathrm{Norm}(\cdot)$ denotes $\ell_2$-normalization, and all normalization operations use a small $\epsilon$ for numerical stability.
The resulting prototypes provide an initialization for adaptation.

\section{Proposed Method}
\sectionnavtarget{nav:method}

\begin{figure}
\centering
\includegraphics[width=\linewidth]{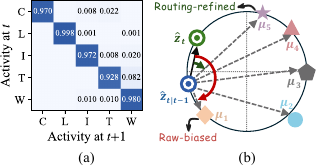}
\caption{\textbf{Motivation of \method.} 
(a)~Output-space transition statistics show strong diagonal dominance, reflecting activity persistence and limited transition flexibility. 
(b)~Feature-space deviation offers an observation-conditioned direction for routing refinement toward aligned activity prototypes.}
\label{fig:motivation}
\end{figure}

\paragraph{Motivation.}
Following Definition~\ref{def:temporal_structure}, WHAR streams are characterized by prolonged activity persistence punctuated by sparse transitions. To leverage this structure, a TTA method for WHAR is implicitly answering two questions at every incoming window: \emph{when} the model should stay with its recent belief, and once it should not, \emph{where} the prediction should move. A natural way to encode these two questions together is a Markov transition matrix over activity labels~\citep{HybridDiscriminativeGenerative_IJCAI2005,HARMarkov_CIN2011,OnlineMarkov-HAR_JAIHC2020}, whose diagonal governs when to stay and whose off-diagonals govern where to go. This output-space view has recently been carried into TTA as a transition prior for streaming inference~\citep{OATTA_arXiv2026}. Yet such a matrix resides entirely in the output space and is estimated from past predictions, so it is vulnerable to noisy historical predictions and often suffers from strong diagonal dominance, as shown in Figure~\ref{fig:motivation}~(a). It is moreover blind to the current observation: it over-trusts past persistence and routes transitions purely by historical label co-occurrence, producing an \emph{inertial deadlock} that obstructs timely adaptation to emerging activities. This pushes us to step out of the output space and revisit temporal structure through the lens of feature-space geometry. The geometric gap between the expected feature state under persistence and the observed feature serves as a per-window cue for \emph{when} the recent belief no longer fits. Crucially, this gap is directional rather than scalar, and its orientation in feature space can indicate \emph{where} the activity is drifting toward, grounding transition routing in the current observation rather than in transition-frequency statistics, as illustrated in Figure~\ref{fig:motivation}~(b).

\begin{figure}
\centering
\includegraphics[width=\linewidth]{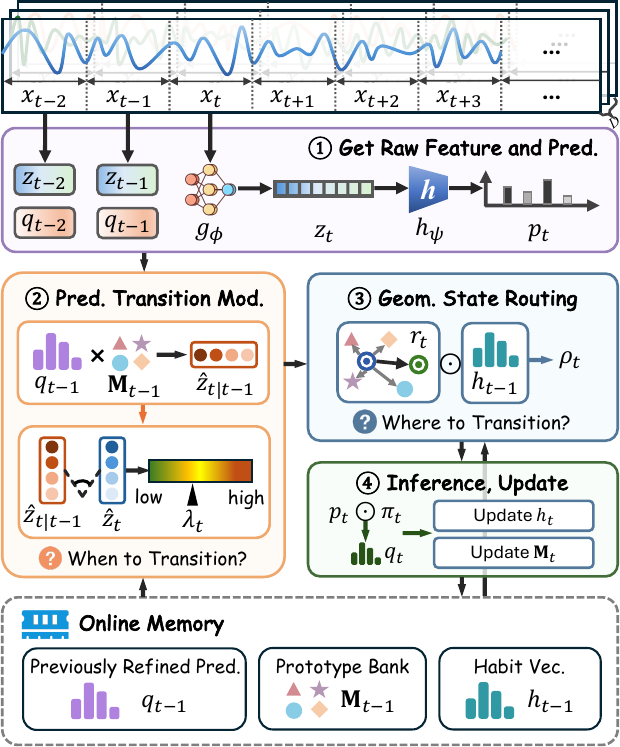}
\caption{\textbf{Overview of \method.} It has predictive transition modulation, geometric state routing, and memory update.}
\label{fig:framework}
\end{figure}

\paragraph{Framework Overview.}
We propose \textbf{\method}, an efficient TTA framework for WHAR. \method initializes class prototypes from the source classifier weights as geometric anchors. For each target window, it compares the current feature with the feature state expected from the previous refined prediction, producing a predictive surprise signal that adaptively preserves or releases temporal inertia. When a transition is likely, the feature deviation direction routes predictions toward geometrically aligned activity prototypes, calibrated by habit statistics. The resulting prior is fused with the raw classifier output for prediction refinement, while prototypes and the marginal habit vector are updated online to track gradual target stream shifts, without revisiting past samples or backpropagating through the model. Figure~\ref{fig:framework} illustrates the overall framework and its workflow.

\subsection{Predictive Transition Modulation}

Predictive transition modulation aims to make temporal continuity adaptive rather than unconditional. Instead of treating every adjacent window as equally persistent, \method estimates whether the incoming observation remains compatible with recent history. The resulting predictive geometric surprise provides a sample-wise signal for distinguishing ordinary within-activity variation from likely state transitions, thereby relaxing temporal inertia only when the current observation sufficiently violates temporal persistence.

\paragraph{Expected State Projection.}
Predictive transition modulation requires a reference feature describing what the current observation should look like if the previous activity state persists. A simple choice is to use $\hat{z}_{t-1}$ as this reference. However, $\hat{z}_{t-1}$ is only a noisy observation and may vary within the same activity due to motion phase, sensor noise, or window-level sampling effects. Directly comparing $\hat{z}_t$ with $\hat{z}_{t-1}$ therefore captures local fluctuation rather than the expected activity state from recent history.

To obtain a more semantic and belief-conditioned reference, \method projects the previous refined prediction onto the current prototype geometry. The refined prediction $q_{t-1}$ summarizes the previous activity belief, while the prototype bank $\mathbf{M}_{t-1}$ provides class-level feature anchors adapted to the target stream. Their projection gives the feature state expected under temporal persistence. Let $\mathbf{M}_{t-1}=[\mu_1^{(t-1)},\ldots,\mu_K^{(t-1)}]^\top \in \mathbb{R}^{K\times d}$ denote the prototype bank at time $t-1$. Given the previous refined prediction $q_{t-1}\in\Delta^{K-1}$, the expected state is computed as:
\begin{equation}\label{eq:expected-state}
\resizebox{0.909\linewidth}{!}{$
    \hat{z}_{t|t-1}
    = \operatorname{Norm}\!\left(\mathbf{M}_{t-1}^{\top} q_{t-1}\right)
    = \operatorname{Norm}\!\left(\sum_{k=1}^{K} q_{t-1}^{(k)} \mu_k^{(t-1)}\right).
$}
\end{equation}
Here, $\hat{z}_{t|t-1}$ denotes the expected feature state under temporal persistence, while $\hat{z}_t = \operatorname{Norm}(z_t)$ denotes the normalized current observation feature.

\paragraph{Transition Surprise Estimation.}
Given the expected feature state $\hat{z}_{t|t-1}$ and the current observation $\hat{z}_t$, we estimate how strongly the current feature violates temporal persistence.
Since both vectors are $\ell_2$-normalized, their angular discrepancy can be measured by the cosine distance:
\begin{equation}
    \mathcal{D}_t
    =
    1 - \hat{z}_t^\top \hat{z}_{t|t-1}.
\end{equation}
A small $\mathcal{D}_t$ indicates that the incoming feature remains consistent with the state predicted from recent history, whereas a large $\mathcal{D}_t$ suggests a potential activity transition.
We convert this discrepancy into a bounded transition surprise score:
\begin{equation}\label{eq:gate}
    \lambda_t
    =
    1 - \exp\left(-\beta \mathcal{D}_t^2\right),
\end{equation}
where $\beta>0$ controls the sensitivity to feature deviation.
The resulting $\lambda_t\in[0,1]$ serves as a sample-wise inertia release factor:
when the observed feature agrees with the expected state, $\lambda_t$ remains small and temporal inertia is preserved;
when the feature deviates sharply, $\lambda_t$ increases and allows the prediction to move away from the previous state.

\subsection{Geometric--Habit State Routing}

A large surprise score indicates that the previous state may no longer be reliable, but does not by itself specify which activity is emerging in the target stream. Therefore, after temporal inertia is relaxed, \method must determine where the prediction should move, beyond Markovian label-frequency priors that ignore the current observation. To resolve this ambiguity, \method performs geometric attentional routing by comparing the observation-induced deviation with directions toward candidate activity prototypes, and complements this local geometric evidence with a marginal-habit calibration prior to incorporate stream-level activity preferences.

\paragraph{Geometric Vector Attention.}
To make transition routing conditioned on the current observation, we compare the direction of the observed feature movement with the directions toward candidate activity prototypes.
Specifically, we define the normalized observation-induced displacement as:
\begin{equation}
    v_t
    =
    \operatorname{Norm}\!\left(\hat{z}_t-\hat{z}_{t|t-1}\right),
\end{equation}
where $\hat{z}_{t|t-1}$ is the expected feature state and $\hat{z}_t$ is the normalized observed feature.
For each candidate activity $k$, we define its prototype direction from the expected state as:
\begin{equation}
    u_{t,k}
    =
    \operatorname{Norm}\!\left(\mu_k^{(t-1)}-\hat{z}_{t|t-1}\right).
\end{equation}
We then measure how well each candidate direction aligns with the observed feature movement:
\begin{equation}
    a_{t,k}
    =
    v_t^\top u_{t,k}.
\end{equation}
The alignment scores are converted into an attentional routing distribution:
\begin{equation}
    r_{t,k}
    =
    \frac{\exp(a_{t,k}/\tau)}
    {\sum_{j=1}^{K}\exp(a_{t,j}/\tau)},
    \quad k=1,\ldots,K,
\end{equation}
where $\tau$ controls the sharpness of the geometric attention distribution.
The resulting $r_t\in\Delta^{K-1}$ serves as an observation-conditioned routing prior, assigning higher probability to candidate states whose prototype directions are better aligned with the current feature displacement.

\paragraph{Marginal-Habit Prior Calibration.}
The geometric routing distribution $r_t$ captures the transition tendency induced by the current feature movement, but it does not account for stream-level activity preferences accumulated over time.
To incorporate such marginal context while mitigating the dominance of frequent activities, we maintain a habit vector $h_{t-1}\in\Delta^{K-1}$ and apply sub-linear flattening:
\begin{equation}
    \tilde{h}_{t-1}
    =
    \Pi_\Delta\!\left(\sqrt{h_{t-1}+\epsilon}\right),
\end{equation}
where $\epsilon$ is a small constant for numerical stability. Compared with directly using $h_{t-1}$, the square-root transform preserves the relative ordering of activity frequencies while reducing the dominance of high-frequency states.
We then calibrate the geometric routing prior by combining it with the flattened marginal prior:
\begin{equation}
    \rho_t
    =
    \Pi_\Delta\!\left(r_t \odot \tilde{h}_{t-1}\right),
\end{equation}
where $\Pi_\Delta(a)=a/\sum_k a_k$ normalizes a non-negative vector onto the probability simplex, and $\odot$ denotes element-wise multiplication.
The resulting $\rho_t$ balances local feature-driven transition evidence with stream-level marginal prevalence, providing a calibrated prior over candidate activity states.

\subsection{Online Inference and Update}

With the transition surprise $\lambda_t$ and calibrated routing prior $\rho_t$, \method refines the current prediction and updates its online state. We initialize $h_0$ uniformly, set $q_1=p_1$ for the first window without prior history, and use each $q_t$ as the next prior belief. At subsequent steps, \method integrates observation evidence with temporal structure, then refreshes marginal habit statistics and prototypes for future inference. 

\paragraph{Belief Refinement via Consensus.}
The transition surprise score $\lambda_t$ determines how much the model should move from the previous refined prediction, while the calibrated routing prior $\rho_t$ specifies where the transition should be directed.
We therefore form the temporal prior by interpolating between state persistence and geometry-guided transition:
\begin{equation}
    \pi_t
    =
    (1-\lambda_t) q_{t-1}
    +
    \lambda_t \rho_t.
\end{equation}
Small $\lambda_t$ keeps $\pi_t$ close to $q_{t-1}$ for continuity, whereas larger $\lambda_t$ shifts the prior toward states favored by geometric routing and marginal calibration.
The refined prediction calibrates rather than overrides ambiguous or biased raw outputs, requiring non-negligible class support:
\begin{equation}
    q_t
    =
    \Pi_\Delta\!\left(p_t \odot \pi_t\right).
\end{equation}

\paragraph{Marginal Habit Tracking.}
After obtaining the refined prediction $q_t$, \method updates the habit vector to summarize stream-level marginal activity prevalence.
Instead of committing to hard pseudo-labels, we treat $q_t$ as a soft observation and update the habit state by exponential averaging:
\begin{equation}
    h_t
    =
    (1-\eta_h)h_{t-1}
    +
    \eta_h q_t ,
\end{equation}
where $\eta_h$ controls the tracking speed.
The resulting $h_t$ provides a lightweight memory of marginal activity tendencies for future prior calibration.

\paragraph{Soft Prototype Adaptation.}
To track gradual feature shifts in the target stream, \method updates class prototypes using the refined prediction as a soft assignment, following the common practice of adapting classifier-induced prototypes in source-free settings~\citep{SHOT_ICML2020,ProtoDA_NeurIPS2021,UDASBP_IS2022}.
For each class $k$, we first compute an assignment-weighted prototype update:
\begin{equation}
    \bar{\mu}_k^{(t)}
    =
    \operatorname{Norm}\!\left(
    (1-\eta_\mu q_t^{(k)})\mu_k^{(t-1)}
    +
    \eta_\mu q_t^{(k)}\hat{z}_t
    \right),
\end{equation}
where $\eta_\mu$ controls the prototype adaptation rate.
This soft update avoids committing to noisy hard pseudo-labels, while the factor $q_t^{(k)}$ makes uncertain classes move only mildly toward the current observation.
To further reduce the risk of prototype collapse and long-term drift, we elastically anchor each updated prototype to its source-side initialization:
\begin{equation}
    \mu_k^{(t)}
    =
    \operatorname{Norm}\!\left(
    (1-\omega_\mu)\bar{\mu}_k^{(t)}
    +
    \omega_\mu \mu_k^{(0)}
    \right),
\end{equation}
where $\omega_\mu$ controls the anchoring strength.
Thus, prototypes can adapt to target-domain geometry while constrained by semantic directions learned from the source classifier.

\section{Experiments}\label{sec:exp}
\sectionnavtarget{nav:exp}

We comprehensively evaluate \method through five axes: \textbf{RQ1} (Superiority), \textbf{RQ2} (Effectiveness), \textbf{RQ3} (Sensitivity), \textbf{RQ4} (Efficiency), and \textbf{RQ5} (Adaptability). In the Appendix, we provide additional experimental details and results.

\subsection{Experimental Setup}

\paragraph{Datasets and Preprocessing.}
We use HARTH\footnote{We use the HARTH v2.0 release (2024), as updated on GitHub, which consists of acceleration data from 31 subjects.}~\citep{HARTH_Sensors2021} and CAPTURE-24~\citep{CAPTURE-24_SciData2024}, since they provide continuous real-world, free-living activity streams with preserved temporal structure and natural activity ordering, unlike many conventional WHAR datasets based on scripted activities or pre-segmented samples. We adopt the standard cross-subject evaluation protocol, treating each subject as a distinct domain. We apply a sliding window technique to segment the continuous time-series data into fixed-length segments while maintaining the continuity between segments to preserve temporal dependency and activity transition information. For HARTH, we use a 2-second window with a 1-second stride at 50 Hz, while for CAPTURE-24, we use a 10-second non-overlapping window at 100 Hz. To simulate real-world data streams, we strictly preserve the chronological ordering of test segments during adaptation, avoiding the random shuffling used in conventional protocols. 
For each dataset, we randomly select six distinct source--target pairs for the main study.

\paragraph{Baselines.}
We compare \method with several state-of-the-art methods, including (1)~\textit{vanilla TTA}: TENT~\citep{Tent_ICLR2021}, NOTE~\citep{NOTE_NeurIPS2022}, SAR~\citep{SAR_ICLR2023}, RoTTA~\citep{RoTTA_CVPR2023}, and OATTA~\citep{OATTA_arXiv2026}, (2)~\textit{WHAR-related TTA}: OFTTA~\citep{OFTTA_Ubicomp2024}, ACCUP~\citep{ACCUP_KDD2025} and COA-HAR~\citep{COA-HAR_ESWA2026}. We also report the results of the source-only baseline without test-time adaptation for reference.
\begin{table*}[t]\small
\centering
\renewcommand{\arraystretch}{1.135}
\setlength{\tabcolsep}{1.8mm}
\begin{tabular}{r||cccccc|c}
\hline
\specialrule{.1em}{0em}{0em}
\rowcolor{mygray}
Method & S006 $\rightarrow$ S032 & S008 $\rightarrow$ S033 & S013 $\rightarrow$ S031 & S015 $\rightarrow$ S034 & S016 $\rightarrow$ S038 & S022 $\rightarrow$ S035 & Avg \\
\hline\hline
Source-only & 45.87{\scriptsize $\pm$1.09} $-$ & 52.37{\scriptsize $\pm$0.47} $-$ & 74.83{\scriptsize $\pm$3.06} $-$ & 53.71{\scriptsize $\pm$3.66} $-$ & 96.52{\scriptsize $\pm$0.62} $-$ & 78.25{\scriptsize $\pm$0.94} $-$ & 66.93 $-$ \\
\hhline{-||-|-|-|-|-|-|-}
TENT {\scriptsize {[ICLR'21]}} & 45.92{\scriptsize $\pm$1.01} $\uparrow$ & \underline{56.78}{\scriptsize $\pm$5.93} $\uparrow$ & 73.09{\scriptsize $\pm$3.23} $\downarrow$ & 53.40{\scriptsize $\pm$4.00} $\downarrow$ & 96.28{\scriptsize $\pm$0.55} $\downarrow$ & 85.39{\scriptsize $\pm$9.31} $\uparrow$ & 68.48 $\uparrow$ \\
NOTE {\scriptsize {[NeurIPS'22]}} & 46.23{\scriptsize $\pm$0.75} $\uparrow$ & 52.24{\scriptsize $\pm$0.73} $\downarrow$ & 63.70{\scriptsize $\pm$4.27} $\downarrow$ & 58.37{\scriptsize $\pm$6.82} $\uparrow$ & 56.87{\scriptsize $\pm$7.88} $\downarrow$ & 57.04{\scriptsize $\pm$6.84} $\downarrow$ & 55.74 $\downarrow$ \\
SAR {\scriptsize {[ICLR'23]}} & 39.02{\scriptsize $\pm$3.55} $\downarrow$ & 49.05{\scriptsize $\pm$1.17} $\downarrow$ & 60.06{\scriptsize $\pm$5.55} $\downarrow$ & 55.79{\scriptsize $\pm$3.62} $\uparrow$ & 88.37{\scriptsize $\pm$2.70} $\downarrow$ & 68.53{\scriptsize $\pm$5.22} $\downarrow$ & 60.14 $\downarrow$ \\
RoTTA {\scriptsize {[CVPR'23]}} & 45.87{\scriptsize $\pm$1.09} $\uparrow$ & 52.38{\scriptsize $\pm$0.47} $\uparrow$ & 75.12{\scriptsize $\pm$3.14} $\uparrow$ & 53.90{\scriptsize $\pm$3.49} $\uparrow$ & \underline{96.67}{\scriptsize $\pm$0.45} $\uparrow$ & 78.24{\scriptsize $\pm$0.94} $\downarrow$ & 67.03 $\uparrow$ \\
OATTA {\scriptsize {[arXiv'26]}} & \underline{48.31}{\scriptsize $\pm$3.34} $\uparrow$ & 56.77{\scriptsize $\pm$4.13} $\uparrow$ & \underline{75.29}{\scriptsize $\pm$3.12} $\uparrow$ & \textbf{62.13}{\scriptsize $\pm$2.85} $\uparrow$ & 96.58{\scriptsize $\pm$0.49} $\uparrow$ & \underline{87.19}{\scriptsize $\pm$4.87} $\uparrow$ & \underline{71.05} $\uparrow$ \\
\hhline{-||-|-|-|-|-|-|-}
OFTTA {\scriptsize {[Ubicomp'24]}} & 45.88{\scriptsize $\pm$1.16} $\uparrow$ & 52.38{\scriptsize $\pm$0.43} $\uparrow$ & 75.20{\scriptsize $\pm$3.05} $\uparrow$ & 54.82{\scriptsize $\pm$3.31} $\uparrow$ & 96.36{\scriptsize $\pm$0.66} $\downarrow$ & 78.13{\scriptsize $\pm$1.17} $\downarrow$ & 67.13 $\uparrow$ \\
ACCUP {\scriptsize {[KDD'25]}} & 21.50{\scriptsize $\pm$4.87} $\downarrow$ & 29.40{\scriptsize $\pm$4.72} $\downarrow$ & 21.59{\scriptsize $\pm$2.49} $\downarrow$ & 15.62{\scriptsize $\pm$1.12} $\downarrow$ & 25.21{\scriptsize $\pm$2.38} $\downarrow$ & 28.74{\scriptsize $\pm$1.86} $\downarrow$ & 23.68 $\downarrow$ \\
COA-HAR {\scriptsize {[ESWA'26]}} & 36.77{\scriptsize $\pm$1.86} $\downarrow$ & 50.13{\scriptsize $\pm$1.29} $\downarrow$ & 47.29{\scriptsize $\pm$0.55} $\downarrow$ & 59.37{\scriptsize $\pm$3.94} $\uparrow$ & 46.09{\scriptsize $\pm$0.99} $\downarrow$ & 43.36{\scriptsize $\pm$6.21} $\downarrow$ & 47.17 $\downarrow$ \\
\hhline{-||-|-|-|-|-|-|-}
\textbf{\method} (ours) & \textbf{58.27}{\scriptsize $\pm$1.19} $\uparrow$ & \textbf{65.72}{\scriptsize $\pm$0.51} $\uparrow$ & \textbf{79.17}{\scriptsize $\pm$2.99} $\uparrow$ & \underline{61.82}{\scriptsize $\pm$2.94} $\uparrow$ & \textbf{97.02}{\scriptsize $\pm$0.51} $\uparrow$ & \textbf{98.36}{\scriptsize $\pm$0.03} $\uparrow$ & \textbf{76.73} $\uparrow$ \\
\hline 
\specialrule{.1em}{0em}{0em}
\end{tabular}
\caption{MF1-scores (\%) of \method and the baselines on HARTH dataset.}
\label{tab:exp:main_HARTH}
\end{table*}

\begin{table*}[t]\small
\centering
\renewcommand{\arraystretch}{1.135}
\setlength{\tabcolsep}{1.8mm}
\begin{tabular}{r||cccccc|c}
\hline
\specialrule{.1em}{0em}{0em}
\rowcolor{mygray}
Method & P002 $\rightarrow$ P025 & P004 $\rightarrow$ P036 & P007 $\rightarrow$ P021 & P009 $\rightarrow$ P034 & P012 $\rightarrow$ P026 & P013 $\rightarrow$ P029 & Avg \\
\hline\hline
Source-only & 42.69{\scriptsize $\pm$1.18} $-$ & 39.78{\scriptsize $\pm$1.06} $-$ & 56.76{\scriptsize $\pm$1.05} $-$ & 37.31{\scriptsize $\pm$1.50} $-$ & 47.66{\scriptsize $\pm$1.85} $-$ & 24.29{\scriptsize $\pm$1.91} $-$ & 41.42 $-$ \\
\hhline{-||-|-|-|-|-|-|-}
TENT {\scriptsize {[ICLR'21]}} & 42.64{\scriptsize $\pm$1.71} $\downarrow$ & 40.13{\scriptsize $\pm$0.87} $\uparrow$ & 56.64{\scriptsize $\pm$1.50} $\downarrow$ & 37.09{\scriptsize $\pm$1.69} $\downarrow$ & 47.51{\scriptsize $\pm$1.83} $\downarrow$ & 23.69{\scriptsize $\pm$2.36} $\downarrow$ & 41.28 $\downarrow$ \\
NOTE {\scriptsize {[NeurIPS'22]}} & 33.85{\scriptsize $\pm$8.58} $\downarrow$ & 29.40{\scriptsize $\pm$1.26} $\downarrow$ & 58.41{\scriptsize $\pm$1.96} $\uparrow$ & 35.42{\scriptsize $\pm$1.28} $\downarrow$ & \underline{51.75}{\scriptsize $\pm$1.43} $\uparrow$ & 18.20{\scriptsize $\pm$3.04} $\downarrow$ & 37.84 $\downarrow$ \\
SAR {\scriptsize {[ICLR'23]}} & 23.91{\scriptsize $\pm$1.47} $\downarrow$ & 25.90{\scriptsize $\pm$0.82} $\downarrow$ & 34.32{\scriptsize $\pm$2.85} $\downarrow$ & 32.14{\scriptsize $\pm$1.42} $\downarrow$ & 33.11{\scriptsize $\pm$2.05} $\downarrow$ & 23.13{\scriptsize $\pm$1.59} $\downarrow$ & 28.75 $\downarrow$ \\
RoTTA {\scriptsize {[CVPR'23]}} & 42.53{\scriptsize $\pm$1.16} $\downarrow$ & 39.74{\scriptsize $\pm$1.08} $\downarrow$ & 56.64{\scriptsize $\pm$1.15} $\downarrow$ & 37.27{\scriptsize $\pm$1.57} $\downarrow$ & 45.25{\scriptsize $\pm$1.34} $\downarrow$ & 24.18{\scriptsize $\pm$1.78} $\downarrow$ & 40.94 $\downarrow$ \\
OATTA {\scriptsize {[arXiv'26]}} & \underline{46.85}{\scriptsize $\pm$1.23} $\uparrow$ & \underline{42.62}{\scriptsize $\pm$1.19} $\uparrow$ & \underline{61.93}{\scriptsize $\pm$1.83} $\uparrow$ & 37.50{\scriptsize $\pm$1.90} $\uparrow$ & 50.81{\scriptsize $\pm$0.72} $\uparrow$ & \underline{27.69}{\scriptsize $\pm$4.35} $\uparrow$ & \underline{44.57} $\uparrow$ \\
\hhline{-||-|-|-|-|-|-|-}
OFTTA {\scriptsize {[Ubicomp'24]}} & 42.61{\scriptsize $\pm$1.64} $\downarrow$ & 39.54{\scriptsize $\pm$0.86} $\downarrow$ & 56.26{\scriptsize $\pm$1.38} $\downarrow$ & \underline{38.06}{\scriptsize $\pm$2.06} $\uparrow$ & 45.74{\scriptsize $\pm$0.94} $\downarrow$ & 25.73{\scriptsize $\pm$2.67} $\uparrow$ & 41.32 $\downarrow$ \\
ACCUP {\scriptsize {[KDD'25]}} & 21.35{\scriptsize $\pm$0.18} $\downarrow$ & 13.95{\scriptsize $\pm$1.10} $\downarrow$ & 18.93{\scriptsize $\pm$6.32} $\downarrow$ & 20.14{\scriptsize $\pm$4.32} $\downarrow$ & 12.65{\scriptsize $\pm$2.79} $\downarrow$ & 23.17{\scriptsize $\pm$3.51} $\downarrow$ & 18.37 $\downarrow$ \\
COA-HAR {\scriptsize {[ESWA'26]}} & 27.12{\scriptsize $\pm$2.79} $\downarrow$ & 22.29{\scriptsize $\pm$1.02} $\downarrow$ & 43.41{\scriptsize $\pm$3.10} $\downarrow$ & 31.65{\scriptsize $\pm$1.44} $\downarrow$ & 47.09{\scriptsize $\pm$1.91} $\downarrow$ & 19.29{\scriptsize $\pm$0.89} $\downarrow$ & 31.81 $\downarrow$ \\
\hhline{-||-|-|-|-|-|-|-}
\textbf{\method} (ours) & \textbf{51.49}{\scriptsize $\pm$1.33} $\uparrow$ & \textbf{45.23}{\scriptsize $\pm$2.11} $\uparrow$ & \textbf{67.99}{\scriptsize $\pm$4.02} $\uparrow$ & \textbf{40.06}{\scriptsize $\pm$1.17} $\uparrow$ & \textbf{53.99}{\scriptsize $\pm$1.08} $\uparrow$ & \textbf{33.83}{\scriptsize $\pm$3.68} $\uparrow$ & \textbf{48.77} $\uparrow$ \\
\hline
\specialrule{.1em}{0em}{0em}
\end{tabular}
\caption{MF1-scores (\%) of \method and the baselines on CAPTURE-24 dataset.}
\label{tab:exp:main_CAPTURE-24}
\end{table*}

\paragraph{Implementation Details.}

We adopt the widely-used 1D-CNN architecture as the backbone, and a simple linear layer as the classification head across all the baselines.
The source-trained models are pretrained for 100 epochs using the Adam optimizer with a learning rate of 0.001 and a batch size of 64.
During TTA, we set the learning rate to 0.0001 for the methods that require parameter updates, and for \method, we set $\beta=1$, $\tau=0.05$, $\eta_\mu=0.005$, $\eta_h=0.05$, and $\omega_\mu=0.01$.
For each source-target pair, we repeat the experiments three times with different random seeds. To ensure a fair comparison, all methods share the exact same set of three pre-trained source models.
Following previous works~\citep{CA-TCC_TPAMI2023,COA-HAR_ESWA2026}, we adopt the \emph{macro-averaged F1-score} (MF1-score) to better evaluate the performance on the imbalanced WHAR datasets.

\subsection{Comparison with State-of-the-Arts (RQ1)}

We benchmark \method against the state-of-the-art methods on HARTH and CAPTURE-24. The comprehensive results are summarized in Table~\ref{tab:exp:main_HARTH} and Table~\ref{tab:exp:main_CAPTURE-24}, respectively.

\paragraph{Performance on HARTH.}
\method achieves the best average performance on HARTH, reaching an MF1-score of 76.73\% and surpassing the strongest baseline by 5.68 percentage points. Many conventional TTA methods even degrade below Source-only under cross-subject shifts, indicating that direct test-time optimization without accounting for temporal structure may amplify the negative effects. In contrast, the temporal dynamics-aware method OATTA is more competitive and slightly leads on one pair, while \method achieves the most reliable overall improvement.

\paragraph{Performance on CAPTURE-24.}
\method also achieves the best performance on the more challenging CAPTURE-24 dataset, where all methods obtain lower absolute MF1-scores due to stronger free-living variability. It reaches an average MF1-score of 48.77\%, improving over Source-only by 7.35 percentage points and the strongest baseline by 4.20 percentage points. The second-best results are distributed across different baselines, suggesting that CAPTURE-24 presents more diverse adaptation difficulties, while \method maintains stable gains across the evaluated transfers.

\subsection{Diagnostic Analysis}

\paragraph{Ablation Study (RQ2).}

We evaluate four mechanisms in \method: (i)~Predictive surprise, (ii)~Geometric Routing, (iii)~Habit Prior, and (iv)~Prototype Update. The ablation results are summarized in Table~\ref{tab:exp:diagnostic_ablation}.
Overall, every variant underperforms the full model, indicating that no module is redundant. The largest drops also differ across datasets, with CAPTURE-24 relying more on prototype adaptation. The consistent drops across datasets suggest that the gains come from complementary temporal, geometric, and habit-aware adaptation signals rather than dataset-specific tuning. In particular, removing transition- or geometry-related cues makes the model more vulnerable to confusing activity changes with local feature noise, while removing habit or prototype updates weakens stream-specific calibration.
\begin{table}[htbp]\small
\centering
\renewcommand{\arraystretch}{1.12}
\setlength{\tabcolsep}{1.4mm}
\begin{tabular}{r||P{1.95cm}P{1.95cm}}
\hline
\specialrule{.1em}{0em}{0em}
\rowcolor{mygray}
Mechanism & HARTH & CAPTURE-24 \\
\hline\hline
Source-only & 66.93 & 41.42 \\
\hhline{-||--}
\textit{w/o} Pred. Surp. & 73.60 & 47.91 \\
\textit{w/o} Geom. Rtg. & 74.19 & 46.87 \\
\textit{w/o} Hab. Prior & 73.78 & 47.75 \\
\textit{w/o} Proto. Upd. & 74.92 & 46.45 \\
\hhline{-||--}
\textbf{Full} & \textbf{76.73} & \textbf{48.77} \\
\hline
\specialrule{.1em}{0em}{0em}
\end{tabular}
\caption{Ablation experimental MF1-scores (\%) of each mechanism of \method on HARTH and CAPTURE-24.}
\label{tab:exp:diagnostic_ablation}
\end{table}

\begin{figure}[htbp]
\centering
\includegraphics[width=\linewidth]{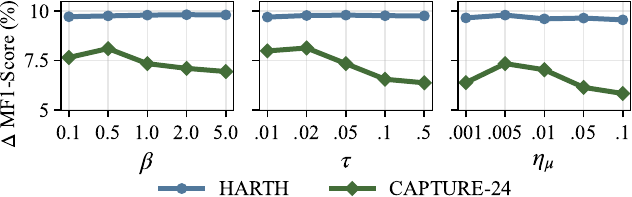}
\caption{\textbf{Sensitivity Analysis.} $\Delta$ MF1-score over Source-only under different parameter values.}
\label{fig:hparams}
\end{figure}

\paragraph{Sensitivity Analysis (RQ3).}
We analyze the sensitivity of \method to three key hyperparameters: (i)~$\boldsymbol{\beta}$ for transition surprise sensitivity, (ii)~$\boldsymbol{\tau}$ for geometric attention sharpness, and (iii)~$\boldsymbol{\eta_\mu}$ for prototype adaptation speed. As shown in Figure~\ref{fig:hparams}, the curves are generally stable across datasets and tested ranges, suggesting that \method is not highly sensitive to hyperparameter choices. HARTH shows relatively flat curves and favors larger $\beta$, suggesting that structured activity streams tolerate broader and more responsive inertia release. A more responsive gate can clarify transitions, whereas noisier free-living deviations require avoiding overreaction. CAPTURE-24 is relatively more sensitive, especially to larger $\beta$, $\tau$, and $\eta_\mu$, whose effects can be amplified by free-living variability, over-smoothing geometric routing or updating prototypes too aggressively. Overall, moderate values provide a better balance between preserving stable activity segments and responding to genuine transitions.

\begin{figure}[htbp]
\includegraphics[width=\linewidth]{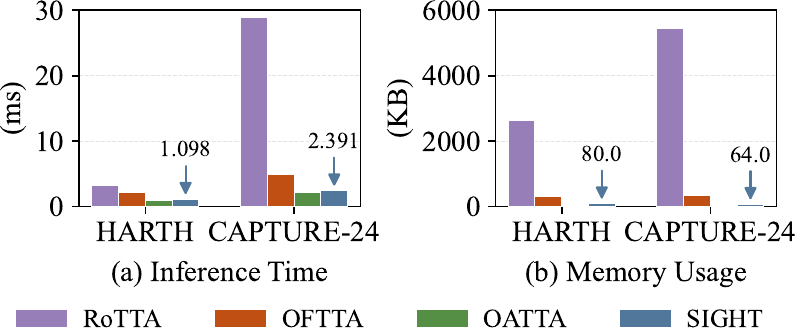}
\centering
\caption{\textbf{Efficiency comparison}. (a)~Per-sample inference time (ms), measured on a single \textbf{CPU} thread. (b)~Memory overhead (KB), defined as the exact byte size of additional tensor state each method stores beyond the shared backbone.}
\label{fig:exp:efficiency}
\end{figure}

\paragraph{Efficiency Analysis (RQ4).}
As WHAR models are typically deployed on resource-constrained edge devices, we evaluate \method in inference time and memory usage. 
Memory-wise, \method only maintains a prototype bank and a habit vector, with memory cost $\mathcal{O}(Kd)$ for $K$ activity classes. Computationally, its surprise estimation, geometric routing, and prototype updates add simple $\mathcal{O}(Kd)$ vector operations per inference step. Since $K$ is usually small in WHAR and no gradients or optimizer states are required, the overhead is minor compared with the model forward pass, making \method suitable for streaming adaptation. We empirically measure the per-sample inference time and memory overhead of \method and the baselines, 
as shown in Figure~\ref{fig:exp:efficiency}. \method achieves inference time comparable to output-space-only OATTA, while being significantly faster than other methods. Its memory overhead is also minimal overall, while some baselines require additional optimizer states.

\paragraph{Adaptability to Backbone Architectures (RQ5).}

\method can be seamlessly applied to various deep model architectures. As Table~\ref{tab:exp:diagnostic_encoder} shows, it consistently improves recognition performance across three distinct backbones: (i)~{1D-CNN}, (ii)~{1D-ResNet}, and (iii)~{Transformer}. These consistent improvements confirm its broad architectural adaptability, while the varying gains suggest that backbone selection should still consider dataset characteristics. 

\begin{table}[htbp]\small
\centering
\renewcommand{\arraystretch}{1.12}
\setlength{\tabcolsep}{1.05mm}
\begin{tabular}{c|r||ccc}
\hline
\specialrule{.1em}{0em}{0em}
\rowcolor{mygray}
Dataset & Method & 1D-CNN & 1D-ResNet & Transformer  \\
\hline\hline
 & Src.-only & 66.93 & 67.50 & 67.27 \\
\hhline{*1>{\arrayrulecolor{white}}->{\arrayrulecolor{black}}|-||---}
 & RoTTA & 67.03 & 67.47 & 67.18 \\
 & OATTA & 71.05 & 70.22 & 68.91 \\
 & OFTTA & 67.13 & 67.20 & 66.96 \\
\multirow{-5}{*}{HARTH} & \textbf{\method} & \textbf{76.73} & \textbf{74.98} & \textbf{73.19} \\
\hhline{-|-||---}
 & Src.-only & 41.42 & 40.56 & 41.08 \\
\hhline{*1>{\arrayrulecolor{white}}->{\arrayrulecolor{black}}|-||---}
 & RoTTA & 40.94 & 40.46 & 42.08 \\
 & OATTA & 44.57 & 44.11 & 44.93 \\
 & OFTTA & 41.32 & 40.53 & 40.57 \\
\multirow{-5}{*}{CAPTURE-24} & \textbf{\method} & \textbf{48.77} & \textbf{47.59} & \textbf{46.85} \\
\hline
\specialrule{.1em}{0em}{0em}
\end{tabular}
\caption{Average MF1-scores (\%) of \method and the baselines on HARTH and CAPTURE-24.}
\label{tab:exp:diagnostic_encoder}
\end{table}

\section{Conclusion}\label{sec:conclusion}
\sectionnavtarget{nav:conclusion}
In this paper, we studied test-time adaptation for wearable human activity recognition under cross-subject shifts. We revisited activity persistence and transitions as feature-conditioned signals rather than output-space smoothing priors, and proposed \method, a lightweight and backpropagation-free framework for online prediction refinement. By estimating predictive surprise and routing adaptation with feature geometry and stream-level habits, \method balances temporal stability with timely response to activity changes. Experiments on HARTH and CAPTURE-24 show that \method achieves superior overall performance over existing TTA baselines, while diagnostic studies verify its component contributions and adaptability across backbones. These results highlight conditioning temporal inertia on feature evidence, offering a practical direction for efficient WHAR adaptation in streaming deployment.

\bibliography{main}

\clearpage
\setcounter{secnumdepth}{2}

% Set algorithmic keywords for input/output
\renewcommand{\algorithmicrequire}{\textbf{Input:}}
\renewcommand{\algorithmicensure}{\textbf{Output:}}

\appendix
\renewcommand{\thefigure}{A\arabic{figure}}
\renewcommand{\thetable}{A\arabic{table}}
\renewcommand{\theequation}{A\arabic{equation}}
\renewcommand{\thesection}{A\arabic{section}}
\renewcommand{\thesubsection}{\thesection.\arabic{subsection}}
\renewcommand{\thesubsubsection}{\thesubsection.\arabic{subsubsection}}
\renewcommand{\theproposition}{A.\arabic{proposition}}
\renewcommand{\theremark}{A.\arabic{remark}}
\renewcommand{\theproof}{A.\arabic{proof}}
\setcounter{figure}{0}
\setcounter{table}{0}
\setcounter{equation}{0}
\setcounter{proposition}{0}
\setcounter{remark}{0}
\setcounter{proof}{0}

\twocolumn[{%
  \centering
  \vspace{0.15in minus 0.10in}
  {\sffamily\bfseries\fontsize{20}{18}\selectfont Appendix\par}
  \vspace{0.2in plus 0.10in minus 0.08in}
\vspace{0.2in minus 0.10in}
}]

\sectionnavtarget{nav:appendix}
\section{Overall Procedure of \method}

\method operates on the target stream in a single online pass. It first initializes the class prototypes from the source-trained linear classifier weights and sets the habit vector to a uniform distribution, i.e., $h_0 = \mathbf{1}/K$, which provides a neutral stream-level prior before observing target samples. For the first window, \method uses the source model prediction as the refined prediction, i.e., $q_1=p_1$. For each subsequent window, \method projects the previous refined prediction onto the current prototype bank to form an expected feature state, estimates the surprise between this expectation and the current observation, and uses the observed feature displacement to construct a geometry-aware routing prior calibrated by the habit statistics. The resulting temporal prior is fused with the raw prediction to obtain $q_t$, after which both the habit vector and prototypes are updated online for future inference. Algorithm~\ref{app:alg:sight} summarizes the overall procedure.

\begin{algorithm}[h]
\caption{Overall Pipeline of \method}
\label{app:alg:sight}
\begin{algorithmic}[1]
\REQUIRE Source-trained model $f_\theta$, stream $\{x_t\}_{t=1}^{N}$, hyperparameters $\beta,\tau,\eta_\mu,\eta_h,\omega_\mu$;
\ENSURE Refined predictions $\{q_t\}_{t=1}^{N}$;

\STATE Initialize prototypes $\{\mu_k^{(0)}\}_{k=1}^{K}$ from classifier weights;
\STATE Initialize $h_0$ uniformly;

\FOR{$t=1$ to $N$}
    \STATE Obtain feature $\hat{z}_t$ and raw prediction $p_t$;
    \IF{$t=1$}
        \STATE Set $q_t=p_t$;
    \ELSE
        \STATE Project $q_{t-1}$ to expected state $\hat{z}_{t|t-1}$;
        \STATE Estimate surprise $\lambda_t$ from $\hat{z}_t$ and $\hat{z}_{t|t-1}$;
        \STATE Compute geometric routing $r_t$;
        \STATE Calibrate $r_t$ with habit statistics to obtain $\rho_t$;
        \STATE Fuse prior $\pi_t=(1-\lambda_t)q_{t-1}+\lambda_t\rho_t$;
        \STATE Refine prediction $q_t=\Pi_\Delta(p_t\odot\pi_t)$;
    \ENDIF
    \STATE Update habit vector $h_t$ with $q_t$;
    \STATE Update prototypes $\{\mu_k^{(t)}\}_{k=1}^{K}$ with soft assignment;
\ENDFOR

\RETURN $\{q_t\}_{t=1}^{N}$;
\end{algorithmic}
\end{algorithm}

\section{Experiments}

\subsection{Details of Datasets}

We use two real-world and free-living datasets, HARTH~\citep{HARTH_Sensors2021} and CAPTURE-24~\citep{CAPTURE-24_SciData2024}, to evaluate the performance of \method and the baselines. Conventional WHAR datasets, like UCI-HAR~\citep{UCI-HAR_ESANN2013}, are collected either in a controlled lab environment or following a scripted protocol, lacking the naturalness and diversity of real-world scenarios. In contrast, recent datasets like HARTH and CAPTURE-24 are collected in free-living conditions, capturing a wide range of activities and contexts, thus providing a more realistic benchmark for evaluating WHAR models. The details of the two datasets are as follows.
\begin{itemize}
	\item \textbf{HARTH}~\citep{HARTH_Sensors2021} is a free-living WHAR dataset collected using two tri-axial accelerometers placed on the thigh and lower back. We use the latest v2.0 release with 31 subjects and segment the 50 Hz streams into 2-second windows with a 1-second stride.
	\item \textbf{CAPTURE-24}~\citep{CAPTURE-24_SciData2024} is a large-scale in-the-wild dataset collected from 151 participants using wrist-worn tri-axial accelerometers. It contains 3883 hours of acceleration data, of which 2562 hours are annotated, and we use non-overlapping 10-second windows at 100 Hz.
\end{itemize}

HARTH and CAPTURE-24 are representative public datasets collected in free-living conditions without scripted protocols. Nevertheless, we include conventional scripted datasets USC-HAD~\citep{USCHAD_UbiComp2012} and UCI-HAR~\citep{UCI-HAR_ESANN2013} to provide complementary results on the performance of \method and baselines under controlled conditions. The details of the datasets are as follows.

\begin{itemize}
    \item \textbf{USC-HAD}~\citep{USCHAD_UbiComp2012} is an HAR dataset collected from 14 subjects using a waist-mounted inertial sensor, covering 12 daily activities. We segment the 100 Hz streams into 1-second windows with a 0.5-second stride, using tri-axial accelerometer and gyroscope signals as 6 input channels.
    \item \textbf{UCI-HAR}~\citep{UCI-HAR_ESANN2013} is a smartphone-based HAR dataset collected from 30 subjects carrying a waist-mounted smartphone, using accelerometer and gyroscope signals to recognize 6 activities. We use the pre-segmented 2.56-second windows with 50\% overlap at 50 Hz, yielding 6 channels from the tri-axial accelerometer and tri-axial gyroscope.
\end{itemize}

% These two datasets are widely used in the HAR community and provide a useful reference point for evaluating TTA methods under more traditional conditions, where chronological temporal structure is weaker than in free-living streams.
% They also help assess whether the method remains competitive beyond our main free-living setting. We provide detailed results on these datasets in Section~\ref{app:sec:exp:more_results}.

\subsection{Details of Baselines}

In the main text, we use two groups of baselines for comparison: (1)~conventional TTA methods, including TENT~\citep{Tent_ICLR2021}, NOTE~\citep{NOTE_NeurIPS2022}, SAR~\citep{SAR_ICLR2023}, and RoTTA~\citep{RoTTA_CVPR2023}; (2)~recent WHAR-related methods, including OFTTA~\citep{OFTTA_Ubicomp2024}, ACCUP~\citep{ACCUP_KDD2025}, and COA-HAR~\citep{COA-HAR_ESWA2026}. We use the official implementations if available, following their recommended settings when applicable. The details of the baselines are as follows.
\begin{itemize}
	\item \textbf{TENT}~\citep{Tent_ICLR2021} adapts the model by minimizing prediction entropy at test time.
	\item \textbf{NOTE}~\citep{NOTE_NeurIPS2022} improves continual TTA under temporally correlated test samples by maintaining a balanced online adaptation queue.
	\item \textbf{SAR}~\citep{SAR_ICLR2023} performs sharpness-aware and reliable entropy minimization with sample filtering for stable adaptation under dynamic shifts.
	\item \textbf{RoTTA}~\citep{RoTTA_CVPR2023} uses robust batch normalization and a teacher--student mechanism to handle long-term test-time distribution shifts.
	\item \textbf{OATTA}~\citep{OATTA_arXiv2026} performs gradient-free Bayesian refinement with a learned transition matrix and likelihood-ratio gate for weakly structured streams.
	\item \textbf{OFTTA}~\citep{OFTTA_Ubicomp2024} is an optimization-free TTA method for cross-person activity recognition based on adaptive normalization statistics.
	\item \textbf{ACCUP}~\citep{ACCUP_KDD2025} combines augmented contrastive clustering with uncertainty-aware prototype refinement for time-series TTA, which is evaluated on some WHAR datasets in their paper.
	\item \textbf{COA-HAR}~\citep{COA-HAR_ESWA2026} applies contrastive online adaptation with sensor-data augmentation and pseudo-label refinement for WHAR streams.
\end{itemize}

We include additional results of recent TTA methods. The details of these methods are as follows.
\begin{itemize}
    \item \textbf{TCA}~\citep{TCA_ICML2025} aligns test-time feature correlations with a high-confidence pseudo-source bank through lightweight linear transformations.
    \item \textbf{REM}~\citep{REM_ICML2025} stabilizes continual entropy minimization with object masking, masked consistency, and entropy-ranking losses to prevent collapse.
\end{itemize}

\subsection{More Implementation Details}

We provide additional implementation details that complement the experimental setup. The model-specific parameters for 1D-CNN are summarized in Table~\ref{app:tab:implementation_details}. 
For baselines that require mini-batch adaptation, we set the batch size to 32.
The $\epsilon$ for numerical stability is set to $10^{-8}$.
The hyperparameters of \method are determined by grid search on validation pairs disjoint from comparison pairs.

\begin{table}[htbp]
\centering
\renewcommand{\arraystretch}{1.12}
\setlength{\tabcolsep}{4pt}
\begin{tabularx}{\linewidth}{l||X}
\hline
\specialrule{.1em}{0em}{0em}
\rowcolor{mygray}
Parameter & Value \\
\hline
\hline
Input channels & 3 \\
Middle channels & 64 \\
Kernel size & 5 \\
Stride & 1 \\
Dropout rate & 0.1 \\
Final output channels & 128 \\
Feature length & 16 \\
\hline
\specialrule{.1em}{0em}{0em}
\end{tabularx}
\caption{Model-specific implementation parameters.}
\label{app:tab:implementation_details}
\end{table}

\subsection{More Results}\label{app:sec:exp:more_results}

\paragraph{Full Comparison.}
As space is limited in the main text, we provide detailed MF1-scores of \method and all baselines on more source--target pairs in Table~\ref{app:tab:exp:detailed_HARTH} and Table~\ref{app:tab:exp:detailed_CAPTURE-24}. For each dataset, we report 18 additional source--target pairs for a more comprehensive comparison. Overall, the expanded comparison shows a consistent pattern across shifts: \method is more stable than entropy-minimization or prototype-only adaptation methods, and preserves gains across structured HARTH streams and heterogeneous CAPTURE-24. These results suggest that using temporal structure and conservative prototype refinement helps avoid unstable adaptation while exploiting target-stream information.
As a supplementary reference, Table~\ref{app:tab:exp:detailed_USC-HAD} and Table~\ref{app:tab:exp:detailed_UCI-HAR} show that \method also remains competitive on the conventional scripted datasets USC-HAD and UCI-HAR, achieving many best or near-best results, though its advantage is less pronounced. This is likely because these controlled benchmarks exhibit milder shifts and weaker chronological structure due to scripted collection and pre-segmentation, leaving less room for temporal surprise and transition-routing cues to help.

\paragraph{Full Efficiency Results.}
We provide the efficiency comparison of \method and all the baselines in Table~\ref{app:tab:exp:efficiency}, which includes both the inference time and memory usage on HARTH and CAPTURE-24. Notably, compared to optimization-based methods like COA-HAR and RoTTA that require gigabytes of memory, \method maintains a minimal footprint under 100 KB and extremely low latency around 1.1 to 2.4 ms. The results demonstrate that \method achieves a highly favorable balance between performance and efficiency for on-device deployment.

\begin{table}[htbp]\small
\centering
\renewcommand{\arraystretch}{1.12}
\setlength{\tabcolsep}{1.8mm}
\begin{tabular}{r||cc|cc}
\hline
\specialrule{.1em}{0em}{0em}
\rowcolor{mygray}
& \multicolumn{2}{c|}{HARTH} & \multicolumn{2}{c}{CAPTURE-24} \\
\hhline{*1>{\arrayrulecolor{mygray}}->{\arrayrulecolor{black}}||----}
\rowcolor{mygray}
\multirow{-2}{*}{Method} & Time $\downarrow$ & Mem. $\downarrow$ & Time $\downarrow$ & Mem. $\downarrow$ \\
\hline\hline
Source-only & 0.61 & 0.0 & 1.90 & 0.0 \\
\hhline{-||----}
TENT & 3.86 & 821.8 & 9.69 & 813.8 \\
NOTE & 1.83 & 896.8 & 6.86 & 1563.8 \\
SAR & 6.08 & 816.8 & 15.69 & 808.8 \\
RoTTA & 3.23 & 2643.7 & 28.93 & 5431.4 \\
OATTA & \textbf{0.86} & \textbf{0.3} & \textbf{2.20} & \textbf{0.2} \\
OFTTA & 2.23 & 303.7 & 4.86 & 325.0 \\
ACCUP & 9.16 & 1655.8 & 97.94 & 1645.0 \\
COA-HAR & 10.17 & 3986.6 & 74.84 & 6670.6 \\
\hhline{-||----}
\textbf{\method} & \underline{1.10} & \underline{80.0} & \underline{2.39} & \underline{64.0} \\
\hline
\specialrule{.1em}{0em}{0em}
\end{tabular}
\caption{Full efficiency comparison of \method and the baselines (Time: ms, Mem.: KB). Bold and underline indicate the best and second-best results among adaptation methods, excluding Source-only.}
\label{app:tab:exp:efficiency}
\end{table}

\paragraph{Design-Choice Ablation.}

We conduct ablation studies to analyze the contribution of each design choice by replacing it with an alternative. The replacement alternatives are as follows: (i)~for surprise estimation, we replace the proposed geometric surprise with a simple feature distance $\lambda_t=\|\hat{z}_t-\hat{z}_{t|t-1}\|$; (ii)~for habit vector, we replace the Sqrt habit with raw habit; (iii)~for the prototype update, we replace the soft assignment with top-1 hard assignment; and (iv)~also for the prototype update, we remove the source anchoring term. The results are shown in Table~\ref{app:tab:ablation}. Overall, the full \method consistently achieves the best performance on both datasets, confirming the effectiveness of these detailed design choices. The drops are relatively mild on HARTH but more evident on CAPTURE-24, especially for hard prototype assignment and removing source anchoring, suggesting that noisy free-living streams benefit more from conservative prototype adaptation and semantic anchoring.

\begin{table}[htbp]\small
\centering
\renewcommand{\arraystretch}{1.12}
\setlength{\tabcolsep}{2.5mm}
\begin{tabular}{r||cc}
\hline
\specialrule{.1em}{0em}{0em}
\rowcolor{mygray}
Variant & HARTH & CAPTURE-24 \\
\hline\hline
Surprise \textit{w/} Feat. Dist. & 75.82 & 47.80 \\
Habit \textit{w/} Raw Vec. & 75.74 & 48.60 \\
Update \textit{w/} Hard Assign. & 76.21 & 47.91 \\
\textit{w/o} Source Anchoring & 75.98 & 47.73 \\
\hhline{-||--}
\textbf{Full \method} & \textbf{76.73} & \textbf{48.77} \\
\hline
\specialrule{.1em}{0em}{0em}
\end{tabular}
\caption{Ablation experimental MF1-scores (\%) on the detailed design choices of \method.}
\label{app:tab:ablation}
\end{table}

\begin{figure}[htbp]
\centering
\includegraphics[width=0.8\columnwidth]{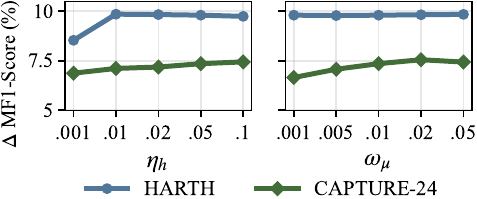}
\caption{\textbf{Sensitivity Analysis.} $\Delta$ MF1-score over Source-only under different parameter ($\eta_h$, $\omega_\mu$) values.}
\label{app:fig:hparams}
\end{figure}

\paragraph{Sensitivity of Other Hyperparameters.}
Besides the main hyperparameters, we also investigate the sensitivity of additional hyperparameters: (i)~$\eta_h$ for habit vector update speed; (ii)~$\omega_\mu$ for prototype anchoring strength. As shown in Figure~\ref{app:fig:hparams}, the curves are generally stable across tested ranges, suggesting that \method is not highly sensitive to these extra hyperparameters. HARTH shows nearly flat performance for both $\eta_h$ and $\omega_\mu$, indicating that structured streams can tolerate a broad range of habit tracking and anchoring strengths. CAPTURE-24 is relatively more sensitive and favors faster habit updates and moderate prototype anchoring, reflecting its stronger free-living variability and greater risk of prototype drift. Overall, moderate values provide a good balance between adapting to target-stream statistics and preserving source-side semantic stability.

\begin{figure}[htbp]
\centering
\includegraphics[width=\columnwidth]{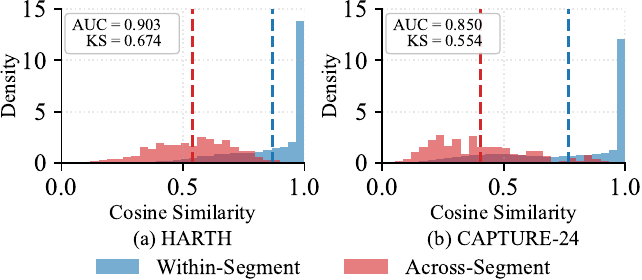}
\caption{\textbf{Transition geometry validation.} Observed vs projected expectation cosine similarities; within-segment and boundary distributions are strongly separable.}
\label{app:fig:transition_geometry}
\end{figure}

\paragraph{Transition Geometry Validation.}
To empirically validate Definition~\ref{def:temporal_structure}~(ii), especially its two geometric assumptions, we extract source features on target streams and compare each observation with its projected expectation under persistence. As shown in Figure~\ref{app:fig:transition_geometry}, the two cosine-similarity distributions are strongly separable, with within-segment similarities concentrated near one and boundary deviations substantially lower, confirming mild variation and pronounced dips. We further assess whether the deviation direction at true transitions is discriminative by scoring candidate classes via alignment with the observed feature shift. The ranking accuracy far exceeds random baselines on both datasets, indicating that transition geometry is informative for online routing. Together, these results verify that boundary deviations are both large enough to trigger surprise and directionally structured enough to guide geometric attention across heterogeneous streams.

\begin{figure}[htbp]
\centering
\includegraphics[width=\columnwidth]{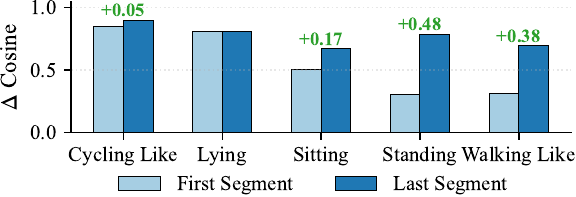}
\caption{\textbf{Prototype alignment improvement.} 
Cosine similarity w.r.t.\ ground-truth centroids for first and last segments. All classes improve.}
\label{app:fig:exp:proto}
\end{figure}

\paragraph{Prototype Refinement Results.}
To verify that \method prototypes accumulate target knowledge, we measure per-segment alignment against ground-truth target centroids using true labels on a pair of HARTH subjects. As elastic anchoring reverts inactive prototypes toward source initialization, step-level curves are dominated by boundary dips. We therefore aggregate per-segment averages and compare the \textbf{first} and \textbf{last} occurrence of each activity in Figure~\ref{app:fig:exp:proto}. Every multi-segment class improves substantially, even from poor initial alignment caused by source classifier weights under severe cross-subject shifts, with large gains observed for classes that were initially far from their ground-truth centroids, confirming that repeated exposure refines prototypes toward target-specific feature geometry. The monotonic increase suggests no catastrophic drift in this analysis.

\paragraph{Chronological-Order Test.}
To quantify each method's reliance on temporal structure, we evaluate under three progressive degradation levels on HARTH and CAPTURE-24. (i)~Chrono.: use the original streaming order. (ii)~Block Perm.: shuffle 32-sample blocks, preserving local continuity but destroying global transitions. (iii)~Full Shfl.: permute windows i.i.d., removing all temporal context. As shown in Table~\ref{app:tab:chrono}, \method achieves its best performance under chronological order and degrades as temporal structure is destroyed, especially on CAPTURE-24. OATTA shows a similar tendency, though with smaller changes, indicating that temporal dynamics-aware methods benefit from ordered activity streams. In contrast, ACCUP and COA-HAR often perform better under block permutation or full shuffling, suggesting that these methods are more aligned with i.i.d. test samples and may be affected by temporally correlated streaming inputs. These results support our motivation that chronological structure is not merely a nuisance in WHAR TTA, but a useful signal when explicitly modeled.

\begin{table}[htbp]\small
\centering
\renewcommand{\arraystretch}{1.12}
\setlength{\tabcolsep}{1.2mm}
% \resizebox{\columnwidth}{!}{%
\begin{tabular}{c|r||ccc}
\hline
\specialrule{.1em}{0em}{0em}
\rowcolor{mygray}
Dataset & Method & Chrono. & Block Perm. & Full Shfl. \\
\hline\hline
 & Src.-only & \multicolumn{3}{c}{All 66.93} \\
 \hhline{*1>{\arrayrulecolor{white}}->{\arrayrulecolor{black}}|-||---}
 & ACCUP & 23.68 & 26.13 & \textbf{29.15} \\
 & COA-HAR & 47.17 & 63.00 & \textbf{63.29} \\
 & OATTA & \textbf{71.05} & 70.93 & 70.97 \\
\multirow{-4}{*}{HARTH} & \textbf{\method} & \textbf{76.73} & 76.18 & 75.12 \\
\hhline{-|-||---}
 & Src.-only & \multicolumn{3}{c}{All 41.42} \\
 \hhline{*1>{\arrayrulecolor{white}}->{\arrayrulecolor{black}}|-||---}
 & ACCUP & 18.37 & 22.23 & \textbf{23.46}  \\
 & COA-HAR & 31.81 & \textbf{38.32} & 37.94 \\
 & OATTA & \textbf{44.57} & 43.99 & 43.04 \\
\multirow{-4}{*}{CAPTURE-24} & \textbf{\method} & \textbf{48.77} & 47.39 & 44.86 \\
\hline
\specialrule{.1em}{0em}{0em}
\end{tabular}
% }
\caption{Macro-F1 on HARTH and CAPTURE-24 under three levels of temporal structure degradation.}
\label{app:tab:chrono}
\end{table}

\section{Discussion}

\paragraph{Role of Temporal Correlation.}
The chronological-order test in Section~\ref{app:sec:exp:more_results} provides a broader view of temporal correlation in WHAR TTA. \method performs best under the original chronological stream and degrades as block permutation and full shuffling progressively remove temporal structure, whereas ACCUP and COA-HAR can improve after shuffling. This contrast suggests that temporal correlation is not merely a non-i.i.d. nuisance or a potential evaluation bias. It can also be an inference signal when modeled explicitly. The key distinction is whether a method can separate ordinary segment persistence from genuine activity transitions. Methods that assume i.i.d.\ batches may benefit from shuffling because it weakens streaming correlation, while \method benefits from chronological order because predictive surprise and geometric routing turn persistence and transitions into usable online evidence.

\paragraph{Why WHAR-Related Baselines Underperform.}
ACCUP~\citep{ACCUP_KDD2025} and COA-HAR~\citep{COA-HAR_ESWA2026} are recent methods designed for time-series TTA, while COA-HAR is specifically proposed for WHAR.
However, both methods degrade severely in our streaming protocol because their designs implicitly assume pre-segmented samples and stable batch statistics rather than the sustained temporal continuity of free-living streams.
They were originally evaluated on traditional scripted HAR datasets.
Such settings often use shuffled pre-segmented samples, which weakens the chronological activity continuity considered in our streaming setting.
ACCUP relies on prototype-based contrastive optimization within each test batch, yet online WHAR batches usually contain only one or two activity classes, making the contrastive signal weak and biased. Cross-subject shift produces noisy early pseudo-labels that contaminate its support sets, while class imbalance leaves rare classes with unreliable centroids. COA-HAR employs periodic batch adaptation with memory replay and InfoNCE contrastive loss, but its contrastive signal requires batch diversity that single-sample streaming cannot offer, while BN updates are destabilized by noisy one-sample statistics. Both methods modify model parameters via backpropagation on streaming pseudo-labels; without the clean target distributions or segmented sample pools their algorithms presuppose, the adaptation can drift rather than improve.

\paragraph{Limitations.}
\method remains subject to several limitations. Although the prototype refinement analysis shows that poorly initialized prototypes can improve through repeated target exposure, early predictions may still be affected by the calibration of the source representation. In addition, \method is most beneficial when the target stream preserves meaningful chronological structure; its advantage may be reduced for heavily shuffled inputs, sparse sampling, or extremely abrupt activity changes where temporal and geometric cues become unreliable. Finally, our experiments focus on accelerometer-based cross-subject WHAR, and extending \method to multi-modal sensing, cross-device shifts, and longer real-world deployments remains an important direction for future work.

\section{Notation}
Table~\ref{app:tab:notation} summarizes the main notation used in the paper.

\begin{table}[htbp]\small
\centering
\renewcommand{\arraystretch}{1.15}
\setlength{\tabcolsep}{4pt}
\begin{tabularx}{\linewidth}{P{1.5cm}||X}
\hline
\specialrule{.1em}{0em}{0em}
\rowcolor{mygray}
Notation & Description \\
\hline
\hline
$f_\theta$ & Source-trained WHAR model. \\
$g_\phi$ & Feature encoder in the model. \\
$h_\psi$ & Linear classifier in the model. \\
$x_t$ & Target sensor window at time $t$. \\
$z_t$ & Raw feature from the encoder. \\
$\hat{z}_t$ & Normalized current feature. \\
$p_t$ & Raw prediction from the source model. \\
$q_t$ & Refined prediction produced by \method. \\
$\mu_k^{(0)}$ & Initial source-side prototype. \\
$\mu_k^{(t)}$ & Online prototype of class $k$. \\
$\mathbf{M}_t$ & Prototype bank at time $t$. \\
$\hat{z}_{t|t-1}$ & Expected feature under temporal persistence. \\
$\mathcal{D}_t$ & Feature discrepancy to expectation. \\
$\lambda_t$ & Surprise score for inertia release. \\
$v_t$ & Observed feature displacement direction. \\
$u_{t,k}$ & Direction toward prototype $k$. \\
$r_t$ & Geometry-based routing distribution. \\
$h_t$ & Stream-level marginal habit vector. \\
$\tilde{h}_t$ & Flattened habit for calibration. \\
$\rho_t$ & Habit-calibrated routing prior. \\
$\pi_t$ & Temporal prior for refinement. \\
$\beta$ & Sensitivity of surprise estimation. \\
$\tau$ & Sharpness of geometric routing. \\
$\eta_h$ & Update rate of habit tracking. \\
$\eta_\mu$ & Update rate of prototypes. \\
$\omega_\mu$ & Strength of source anchoring. \\
\hline
\specialrule{.1em}{0em}{0em}
\end{tabularx}
\caption{Summary of the main notation used in \method.}
\label{app:tab:notation}
\end{table}

\begin{table*}[p]\small
\centering
\renewcommand{\arraystretch}{1.25}
\setlength{\tabcolsep}{1.7mm}
\resizebox{\linewidth}{!}{%
\begin{tabular}{r||cccccccc}
\hline
\specialrule{.1em}{0em}{0em}

\rowcolor{mygray}
Algorithm & S006 $\rightarrow$ S009 & S006 $\rightarrow$ S026 & S008 $\rightarrow$ S015 & S008 $\rightarrow$ S017 & S009 $\rightarrow$ S035 & S013 $\rightarrow$ S023 & S013 $\rightarrow$ S033 & S014 $\rightarrow$ S026 \\
\hline\hline
Source-only & 77.90{\scriptsize $\pm$10.63} $-$ & 74.48{\scriptsize $\pm$8.05} $-$ & 70.69{\scriptsize $\pm$0.64} $-$ & 66.50{\scriptsize $\pm$0.77} $-$ & 35.93{\scriptsize $\pm$0.07} $-$ & 49.54{\scriptsize $\pm$0.26} $-$ & 50.93{\scriptsize $\pm$0.52} $-$ & 41.92{\scriptsize $\pm$7.78} $-$ \\
\hhline{-||-|-|-|-|-|-|-|-}
TENT {\scriptsize {[ICLR'21]}} & 77.93{\scriptsize $\pm$10.76} $\uparrow$ & 73.95{\scriptsize $\pm$8.03} $\downarrow$ & 70.57{\scriptsize $\pm$0.56} $\downarrow$ & 66.70{\scriptsize $\pm$0.85} $\uparrow$ & 35.65{\scriptsize $\pm$0.25} $\downarrow$ & 49.49{\scriptsize $\pm$0.36} $\downarrow$ & 51.32{\scriptsize $\pm$0.71} $\uparrow$ & 45.30{\scriptsize $\pm$8.45} $\uparrow$ \\
NOTE {\scriptsize {[NeurIPS'22]}} & 52.31{\scriptsize $\pm$4.31} $\downarrow$ & 56.76{\scriptsize $\pm$1.44} $\downarrow$ & 66.15{\scriptsize $\pm$2.93} $\downarrow$ & 62.61{\scriptsize $\pm$3.24} $\downarrow$ & 19.65{\scriptsize $\pm$1.56} $\downarrow$ & 49.23{\scriptsize $\pm$0.53} $\downarrow$ & 53.06{\scriptsize $\pm$0.69} $\uparrow$ & 25.71{\scriptsize $\pm$1.38} $\downarrow$ \\
SAR {\scriptsize {[ICLR'23]}} & 31.81{\scriptsize $\pm$0.56} $\downarrow$ & 32.53{\scriptsize $\pm$3.36} $\downarrow$ & 61.40{\scriptsize $\pm$2.82} $\downarrow$ & 58.64{\scriptsize $\pm$1.41} $\downarrow$ & 35.93{\scriptsize $\pm$0.07} $\uparrow$ & \underline{50.00}{\scriptsize $\pm$7.80} $\uparrow$ & 49.95{\scriptsize $\pm$0.76} $\downarrow$ & 18.69{\scriptsize $\pm$3.45} $\downarrow$ \\
RoTTA {\scriptsize {[CVPR'23]}} & 77.90{\scriptsize $\pm$10.63} $\uparrow$ & 74.42{\scriptsize $\pm$8.19} $\downarrow$ & 70.69{\scriptsize $\pm$0.67} $\uparrow$ & 66.52{\scriptsize $\pm$0.76} $\uparrow$ & 35.92{\scriptsize $\pm$0.09} $\downarrow$ & 49.55{\scriptsize $\pm$0.26} $\uparrow$ & 50.95{\scriptsize $\pm$0.53} $\uparrow$ & 42.11{\scriptsize $\pm$7.87} $\uparrow$ \\
OATTA {\scriptsize {[arXiv'26]}} & \underline{89.76}{\scriptsize $\pm$1.05} $\uparrow$ & \underline{80.52}{\scriptsize $\pm$1.98} $\uparrow$ & \underline{72.01}{\scriptsize $\pm$0.87} $\uparrow$ & \underline{68.28}{\scriptsize $\pm$0.57} $\uparrow$ & \underline{36.25}{\scriptsize $\pm$0.09} $\uparrow$ & \underline{50.00}{\scriptsize $\pm$0.23} $\uparrow$ & \underline{58.69}{\scriptsize $\pm$5.52} $\uparrow$ & \underline{46.87}{\scriptsize $\pm$3.75} $\uparrow$ \\
TCA {\scriptsize {[ICML'25]}} & 50.58{\scriptsize $\pm$4.81} $\downarrow$ & 44.66{\scriptsize $\pm$0.55} $\downarrow$ & 37.31{\scriptsize $\pm$4.63} $\downarrow$ & 36.86{\scriptsize $\pm$1.46} $\downarrow$ & 21.68{\scriptsize $\pm$0.72} $\downarrow$ & 29.95{\scriptsize $\pm$3.92} $\downarrow$ & 13.63{\scriptsize $\pm$2.62} $\downarrow$ & 29.59{\scriptsize $\pm$1.55} $\downarrow$ \\
REM {\scriptsize {[ICML'25]}} & 78.00{\scriptsize $\pm$10.72} $\uparrow$ & 74.08{\scriptsize $\pm$8.18} $\downarrow$ & 70.52{\scriptsize $\pm$0.66} $\downarrow$ & 66.71{\scriptsize $\pm$0.78} $\uparrow$ & 35.55{\scriptsize $\pm$0.29} $\downarrow$ & 49.54{\scriptsize $\pm$0.41} $\uparrow$ & 51.38{\scriptsize $\pm$0.73} $\uparrow$ & 44.93{\scriptsize $\pm$8.20} $\uparrow$ \\
OFTTA {\scriptsize {[Ubicomp'24]}} & 78.23{\scriptsize $\pm$10.91} $\uparrow$ & 73.09{\scriptsize $\pm$7.13} $\downarrow$ & 70.69{\scriptsize $\pm$0.56} $\downarrow$ & 66.50{\scriptsize $\pm$0.99} $\uparrow$ & 35.60{\scriptsize $\pm$0.20} $\downarrow$ & 49.67{\scriptsize $\pm$0.13} $\uparrow$ & 50.62{\scriptsize $\pm$0.71} $\downarrow$ & 42.46{\scriptsize $\pm$7.40} $\uparrow$ \\
ACCUP {\scriptsize {[KDD'25]}} & 13.76{\scriptsize $\pm$1.16} $\downarrow$ & 12.18{\scriptsize $\pm$2.62} $\downarrow$ & 24.46{\scriptsize $\pm$3.49} $\downarrow$ & 23.37{\scriptsize $\pm$2.10} $\downarrow$ & 1.98{\scriptsize $\pm$0.34} $\downarrow$ & 23.07{\scriptsize $\pm$1.29} $\downarrow$ & 44.30{\scriptsize $\pm$5.82} $\downarrow$ & 8.84{\scriptsize $\pm$1.67} $\downarrow$ \\
COA-HAR {\scriptsize {[ESWA'26]}} & 39.01{\scriptsize $\pm$8.02} $\downarrow$ & 40.20{\scriptsize $\pm$4.15} $\downarrow$ & 45.73{\scriptsize $\pm$5.21} $\downarrow$ & 57.99{\scriptsize $\pm$3.07} $\downarrow$ & 14.27{\scriptsize $\pm$0.47} $\downarrow$ & 40.84{\scriptsize $\pm$4.36} $\downarrow$ & 51.33{\scriptsize $\pm$1.10} $\uparrow$ & 18.12{\scriptsize $\pm$1.64} $\downarrow$ \\
\textbf{\method} (ours) & \textbf{94.12}{\scriptsize $\pm$1.30} $\uparrow$ & \textbf{83.38}{\scriptsize $\pm$3.00} $\uparrow$ & \textbf{72.17}{\scriptsize $\pm$0.90} $\uparrow$ & \textbf{68.79}{\scriptsize $\pm$0.42} $\uparrow$ & \textbf{38.53}{\scriptsize $\pm$0.05} $\uparrow$ & \textbf{62.87}{\scriptsize $\pm$0.10} $\uparrow$ & \textbf{64.34}{\scriptsize $\pm$0.84} $\uparrow$ & \textbf{49.77}{\scriptsize $\pm$4.00} $\uparrow$ \\

\hline
\specialrule{.1em}{0em}{0em}

\rowcolor{mygray}
Algorithm & S014 $\rightarrow$ S032 & S015 $\rightarrow$ S024 & S015 $\rightarrow$ S028 & S016 $\rightarrow$ S021 & S016 $\rightarrow$ S034 & S018 $\rightarrow$ S030 & S020 $\rightarrow$ S023 & S020 $\rightarrow$ S027 \\
\hline\hline
Source-only & 46.46{\scriptsize $\pm$0.08} $-$ & 74.53{\scriptsize $\pm$0.26} $-$ & 53.33{\scriptsize $\pm$0.01} $-$ & 74.28{\scriptsize $\pm$1.40} $-$ & 55.01{\scriptsize $\pm$0.37} $-$ & 60.63{\scriptsize $\pm$2.02} $-$ & 50.10{\scriptsize $\pm$0.01} $-$ & 46.35{\scriptsize $\pm$1.91} $-$ \\
\hhline{-||-|-|-|-|-|-|-|-}
TENT {\scriptsize {[ICLR'21]}} & \underline{46.57}{\scriptsize $\pm$0.01} $\uparrow$ & 74.52{\scriptsize $\pm$0.30} $\downarrow$ & \textbf{66.67}{\scriptsize $\pm$0.05} $\uparrow$ & 74.33{\scriptsize $\pm$1.56} $\uparrow$ & 55.20{\scriptsize $\pm$0.35} $\uparrow$ & 61.49{\scriptsize $\pm$1.93} $\uparrow$ & 50.11{\scriptsize $\pm$0.01} $\uparrow$ & 50.44{\scriptsize $\pm$3.40} $\uparrow$ \\
NOTE {\scriptsize {[NeurIPS'22]}} & 46.49{\scriptsize $\pm$0.15} $\uparrow$ & 54.55{\scriptsize $\pm$3.95} $\downarrow$ & 58.12{\scriptsize $\pm$4.89} $\uparrow$ & 73.84{\scriptsize $\pm$7.57} $\downarrow$ & 52.45{\scriptsize $\pm$1.94} $\downarrow$ & 34.54{\scriptsize $\pm$4.51} $\downarrow$ & \underline{53.81}{\scriptsize $\pm$3.04} $\uparrow$ & 48.45{\scriptsize $\pm$5.54} $\uparrow$ \\
SAR {\scriptsize {[ICLR'23]}} & 34.85{\scriptsize $\pm$0.53} $\downarrow$ & 76.82{\scriptsize $\pm$6.44} $\uparrow$ & 53.38{\scriptsize $\pm$3.94} $\uparrow$ & 62.21{\scriptsize $\pm$4.09} $\downarrow$ & 50.41{\scriptsize $\pm$1.83} $\downarrow$ & 43.87{\scriptsize $\pm$4.79} $\downarrow$ & 45.75{\scriptsize $\pm$1.09} $\downarrow$ & 37.39{\scriptsize $\pm$1.93} $\downarrow$ \\
RoTTA {\scriptsize {[CVPR'23]}} & 46.49{\scriptsize $\pm$0.09} $\uparrow$ & 74.53{\scriptsize $\pm$0.26} $\uparrow$ & 53.33{\scriptsize $\pm$0.01} $\uparrow$ & 74.19{\scriptsize $\pm$1.32} $\downarrow$ & 55.02{\scriptsize $\pm$0.37} $\uparrow$ & 60.73{\scriptsize $\pm$2.10} $\uparrow$ & 50.11{\scriptsize $\pm$0.01} $\uparrow$ & 46.35{\scriptsize $\pm$1.91} $\uparrow$ \\
OATTA {\scriptsize {[arXiv'26]}} & 46.50{\scriptsize $\pm$0.16} $\uparrow$ & \underline{91.45}{\scriptsize $\pm$8.08} $\uparrow$ & \underline{66.62}{\scriptsize $\pm$0.15} $\uparrow$ & \underline{74.83}{\scriptsize $\pm$1.15} $\uparrow$ & \underline{68.66}{\scriptsize $\pm$0.47} $\uparrow$ & \underline{62.76}{\scriptsize $\pm$2.06} $\uparrow$ & 50.22{\scriptsize $\pm$0.05} $\uparrow$ & 50.18{\scriptsize $\pm$4.54} $\uparrow$ \\
TCA {\scriptsize {[ICML'25]}} & 31.26{\scriptsize $\pm$0.13} $\downarrow$ & 51.78{\scriptsize $\pm$1.80} $\downarrow$ & 30.75{\scriptsize $\pm$2.17} $\downarrow$ & 34.19{\scriptsize $\pm$0.66} $\downarrow$ & 20.60{\scriptsize $\pm$7.44} $\downarrow$ & 26.07{\scriptsize $\pm$2.97} $\downarrow$ & 45.78{\scriptsize $\pm$0.35} $\downarrow$ & 30.88{\scriptsize $\pm$2.59} $\downarrow$ \\
REM {\scriptsize {[ICML'25]}} & 46.56{\scriptsize $\pm$0.02} $\uparrow$ & 74.56{\scriptsize $\pm$0.29} $\uparrow$ & 60.02{\scriptsize $\pm$6.64} $\uparrow$ & 74.26{\scriptsize $\pm$1.55} $\downarrow$ & 55.28{\scriptsize $\pm$0.35} $\uparrow$ & 61.21{\scriptsize $\pm$1.83} $\uparrow$ & 50.12{\scriptsize $\pm$0.02} $\uparrow$ & \underline{50.46}{\scriptsize $\pm$3.38} $\uparrow$ \\
OFTTA {\scriptsize {[Ubicomp'24]}} & 39.63{\scriptsize $\pm$0.47} $\downarrow$ & 74.81{\scriptsize $\pm$0.34} $\uparrow$ & 53.36{\scriptsize $\pm$0.02} $\uparrow$ & 73.66{\scriptsize $\pm$1.19} $\downarrow$ & 54.81{\scriptsize $\pm$0.36} $\downarrow$ & 58.84{\scriptsize $\pm$2.76} $\downarrow$ & 50.02{\scriptsize $\pm$0.08} $\downarrow$ & 47.28{\scriptsize $\pm$1.65} $\uparrow$ \\
ACCUP {\scriptsize {[KDD'25]}} & 31.45{\scriptsize $\pm$0.15} $\downarrow$ & 36.88{\scriptsize $\pm$6.56} $\downarrow$ & 35.43{\scriptsize $\pm$3.02} $\downarrow$ & 20.45{\scriptsize $\pm$2.22} $\downarrow$ & 22.30{\scriptsize $\pm$2.88} $\downarrow$ & 15.32{\scriptsize $\pm$2.45} $\downarrow$ & 16.09{\scriptsize $\pm$2.18} $\downarrow$ & 25.69{\scriptsize $\pm$7.07} $\downarrow$ \\
COA-HAR {\scriptsize {[ESWA'26]}} & 34.45{\scriptsize $\pm$0.09} $\downarrow$ & 50.04{\scriptsize $\pm$3.31} $\downarrow$ & 58.09{\scriptsize $\pm$0.44} $\uparrow$ & 59.49{\scriptsize $\pm$1.66} $\downarrow$ & 47.33{\scriptsize $\pm$1.71} $\downarrow$ & 38.97{\scriptsize $\pm$9.11} $\downarrow$ & 31.11{\scriptsize $\pm$7.57} $\downarrow$ & 33.96{\scriptsize $\pm$3.52} $\downarrow$ \\
\textbf{\method} (ours) & \textbf{58.52}{\scriptsize $\pm$0.20} $\uparrow$ & \textbf{93.18}{\scriptsize $\pm$0.14} $\uparrow$ & 66.38{\scriptsize $\pm$0.05} $\uparrow$ & \textbf{76.20}{\scriptsize $\pm$0.87} $\uparrow$ & \textbf{68.89}{\scriptsize $\pm$0.46} $\uparrow$ & \textbf{67.96}{\scriptsize $\pm$6.74} $\uparrow$ & \textbf{63.30}{\scriptsize $\pm$0.22} $\uparrow$ & \textbf{57.18}{\scriptsize $\pm$1.89} $\uparrow$ \\

\hline
\specialrule{.1em}{0em}{0em}
\end{tabular}
}
\caption{Detailed MF1-scores (\%) of \method and the baselines on additional HARTH source--target pairs.}
\label{app:tab:exp:detailed_HARTH}
\end{table*}

\begin{table*}[p]\small
\centering
\renewcommand{\arraystretch}{1.25}
\setlength{\tabcolsep}{1.7mm}
\resizebox{\linewidth}{!}{%
\begin{tabular}{r||cccccccc}
\hline
\specialrule{.1em}{0em}{0em}

\rowcolor{mygray}
Algorithm & P001 $\rightarrow$ P005 & P003 $\rightarrow$ P006 & P005 $\rightarrow$ P014 & P006 $\rightarrow$ P017 & P007 $\rightarrow$ P018 & P009 $\rightarrow$ P011 & P011 $\rightarrow$ P026 & P013 $\rightarrow$ P021 \\
\hline\hline
Source-only & 48.13{\scriptsize $\pm$3.43} $-$ & 36.34{\scriptsize $\pm$1.50} $-$ & 45.15{\scriptsize $\pm$0.17} $-$ & 47.95{\scriptsize $\pm$0.10} $-$ & 26.82{\scriptsize $\pm$0.18} $-$ & 18.93{\scriptsize $\pm$0.18} $-$ & 42.84{\scriptsize $\pm$0.16} $-$ & 55.88{\scriptsize $\pm$0.30} $-$ \\
\hhline{-||-|-|-|-|-|-|-|-}
TENT {\scriptsize {[ICLR'21]}} & 49.50{\scriptsize $\pm$3.90} $\uparrow$ & 37.52{\scriptsize $\pm$1.59} $\uparrow$ & 45.66{\scriptsize $\pm$0.42} $\uparrow$ & 48.18{\scriptsize $\pm$0.09} $\uparrow$ & 26.76{\scriptsize $\pm$1.61} $\downarrow$ & 18.82{\scriptsize $\pm$0.49} $\downarrow$ & 42.83{\scriptsize $\pm$0.43} $\downarrow$ & 52.99{\scriptsize $\pm$1.46} $\downarrow$ \\
NOTE {\scriptsize {[NeurIPS'22]}} & 44.56{\scriptsize $\pm$3.85} $\downarrow$ & 38.63{\scriptsize $\pm$0.79} $\uparrow$ & 47.51{\scriptsize $\pm$2.34} $\uparrow$ & 35.92{\scriptsize $\pm$3.45} $\downarrow$ & \underline{28.50}{\scriptsize $\pm$0.80} $\uparrow$ & 24.15{\scriptsize $\pm$3.92} $\uparrow$ & 41.05{\scriptsize $\pm$1.24} $\downarrow$ & 52.78{\scriptsize $\pm$2.90} $\downarrow$ \\
SAR {\scriptsize {[ICLR'23]}} & 36.83{\scriptsize $\pm$2.52} $\downarrow$ & 35.19{\scriptsize $\pm$0.86} $\downarrow$ & 28.05{\scriptsize $\pm$2.81} $\downarrow$ & 29.80{\scriptsize $\pm$0.35} $\downarrow$ & 22.98{\scriptsize $\pm$1.47} $\downarrow$ & 25.05{\scriptsize $\pm$3.16} $\uparrow$ & 29.98{\scriptsize $\pm$0.92} $\downarrow$ & 40.89{\scriptsize $\pm$8.64} $\downarrow$ \\
RoTTA {\scriptsize {[CVPR'23]}} & 48.07{\scriptsize $\pm$3.34} $\downarrow$ & 36.00{\scriptsize $\pm$1.01} $\downarrow$ & 45.11{\scriptsize $\pm$0.21} $\downarrow$ & 47.88{\scriptsize $\pm$0.15} $\downarrow$ & 26.81{\scriptsize $\pm$0.18} $\downarrow$ & 18.96{\scriptsize $\pm$0.16} $\uparrow$ & 42.80{\scriptsize $\pm$0.39} $\downarrow$ & 55.81{\scriptsize $\pm$0.38} $\downarrow$ \\
OATTA {\scriptsize {[arXiv'26]}} & \underline{49.95}{\scriptsize $\pm$3.58} $\uparrow$ & \underline{38.77}{\scriptsize $\pm$1.77} $\uparrow$ & \underline{47.83}{\scriptsize $\pm$0.21} $\uparrow$ & \underline{50.51}{\scriptsize $\pm$0.26} $\uparrow$ & 26.49{\scriptsize $\pm$1.14} $\downarrow$ & 19.34{\scriptsize $\pm$0.63} $\uparrow$ & \underline{45.70}{\scriptsize $\pm$0.05} $\uparrow$ & \underline{60.63}{\scriptsize $\pm$0.21} $\uparrow$ \\
TCA {\scriptsize {[ICML'25]}} & 34.61{\scriptsize $\pm$1.00} $\downarrow$ & 30.53{\scriptsize $\pm$3.45} $\downarrow$ & 34.94{\scriptsize $\pm$3.65} $\downarrow$ & 43.12{\scriptsize $\pm$2.36} $\downarrow$ & 28.40{\scriptsize $\pm$3.35} $\uparrow$ & 19.27{\scriptsize $\pm$1.37} $\uparrow$ & 36.17{\scriptsize $\pm$0.69} $\downarrow$ & 45.98{\scriptsize $\pm$0.88} $\downarrow$ \\
REM {\scriptsize {[ICML'25]}} & 49.46{\scriptsize $\pm$3.94} $\uparrow$ & 37.31{\scriptsize $\pm$1.58} $\uparrow$ & 45.62{\scriptsize $\pm$0.38} $\uparrow$ & 48.18{\scriptsize $\pm$0.11} $\uparrow$ & 26.97{\scriptsize $\pm$1.48} $\uparrow$ & 18.83{\scriptsize $\pm$0.42} $\downarrow$ & 42.82{\scriptsize $\pm$0.45} $\downarrow$ & 53.04{\scriptsize $\pm$1.43} $\downarrow$ \\
OFTTA {\scriptsize {[Ubicomp'24]}} & 43.70{\scriptsize $\pm$3.77} $\downarrow$ & 35.97{\scriptsize $\pm$2.66} $\downarrow$ & 44.46{\scriptsize $\pm$0.29} $\downarrow$ & 48.73{\scriptsize $\pm$0.59} $\uparrow$ & 27.00{\scriptsize $\pm$0.06} $\uparrow$ & 19.02{\scriptsize $\pm$0.02} $\uparrow$ & 43.24{\scriptsize $\pm$0.53} $\uparrow$ & 55.79{\scriptsize $\pm$1.87} $\downarrow$ \\
ACCUP {\scriptsize {[KDD'25]}} & 15.27{\scriptsize $\pm$2.18} $\downarrow$ & 34.89{\scriptsize $\pm$0.83} $\downarrow$ & 21.04{\scriptsize $\pm$1.19} $\downarrow$ & 32.40{\scriptsize $\pm$1.92} $\downarrow$ & 25.08{\scriptsize $\pm$0.34} $\downarrow$ & \underline{25.11}{\scriptsize $\pm$2.44} $\uparrow$ & 29.05{\scriptsize $\pm$2.98} $\downarrow$ & 36.18{\scriptsize $\pm$0.53} $\downarrow$ \\
COA-HAR {\scriptsize {[ESWA'26]}} & 37.80{\scriptsize $\pm$1.43} $\downarrow$ & 28.95{\scriptsize $\pm$0.79} $\downarrow$ & 34.68{\scriptsize $\pm$0.46} $\downarrow$ & 22.53{\scriptsize $\pm$0.85} $\downarrow$ & 25.82{\scriptsize $\pm$1.70} $\downarrow$ & \textbf{29.01}{\scriptsize $\pm$2.73} $\uparrow$ & 24.41{\scriptsize $\pm$0.68} $\downarrow$ & 38.55{\scriptsize $\pm$0.65} $\downarrow$ \\
\textbf{\method} (ours) & \textbf{52.93}{\scriptsize $\pm$3.20} $\uparrow$ & \textbf{41.34}{\scriptsize $\pm$2.57} $\uparrow$ & \textbf{48.66}{\scriptsize $\pm$0.41} $\uparrow$ & \textbf{53.23}{\scriptsize $\pm$0.11} $\uparrow$ & \textbf{35.20}{\scriptsize $\pm$3.12} $\uparrow$ & 22.47{\scriptsize $\pm$2.19} $\uparrow$ & \textbf{48.49}{\scriptsize $\pm$0.30} $\uparrow$ & \textbf{64.33}{\scriptsize $\pm$0.50} $\uparrow$ \\

\hline
\specialrule{.1em}{0em}{0em}

\rowcolor{mygray}
Algorithm & P014 $\rightarrow$ P034 & P016 $\rightarrow$ P040 & P019 $\rightarrow$ P025 & P023 $\rightarrow$ P035 & P024 $\rightarrow$ P029 & P028 $\rightarrow$ P036 & P031 $\rightarrow$ P032 & P033 $\rightarrow$ P034 \\
\hline\hline
Source-only & 46.44{\scriptsize $\pm$1.36} $-$ & 64.97{\scriptsize $\pm$0.43} $-$ & 51.16{\scriptsize $\pm$1.06} $-$ & 55.99{\scriptsize $\pm$0.24} $-$ & 27.37{\scriptsize $\pm$2.33} $-$ & 44.34{\scriptsize $\pm$0.90} $-$ & 56.93{\scriptsize $\pm$1.51} $-$ & 46.37{\scriptsize $\pm$1.72} $-$ \\
\hhline{-||-|-|-|-|-|-|-|-}
TENT {\scriptsize {[ICLR'21]}} & 46.81{\scriptsize $\pm$1.27} $\uparrow$ & 65.30{\scriptsize $\pm$0.00} $\uparrow$ & 48.91{\scriptsize $\pm$0.55} $\downarrow$ & 55.10{\scriptsize $\pm$0.24} $\downarrow$ & 26.73{\scriptsize $\pm$2.31} $\downarrow$ & 43.97{\scriptsize $\pm$1.47} $\downarrow$ & 58.95{\scriptsize $\pm$3.65} $\uparrow$ & 46.43{\scriptsize $\pm$1.48} $\uparrow$ \\
NOTE {\scriptsize {[NeurIPS'22]}} & \underline{48.90}{\scriptsize $\pm$0.90} $\uparrow$ & 59.73{\scriptsize $\pm$2.72} $\downarrow$ & 40.26{\scriptsize $\pm$1.16} $\downarrow$ & 51.10{\scriptsize $\pm$1.36} $\downarrow$ & \underline{29.07}{\scriptsize $\pm$2.20} $\uparrow$ & 30.98{\scriptsize $\pm$6.92} $\downarrow$ & 48.91{\scriptsize $\pm$1.68} $\downarrow$ & 42.57{\scriptsize $\pm$2.08} $\downarrow$ \\
SAR {\scriptsize {[ICLR'23]}} & 32.24{\scriptsize $\pm$2.08} $\downarrow$ & 37.20{\scriptsize $\pm$1.49} $\downarrow$ & 27.57{\scriptsize $\pm$3.12} $\downarrow$ & 36.41{\scriptsize $\pm$0.48} $\downarrow$ & 28.93{\scriptsize $\pm$1.65} $\uparrow$ & 25.32{\scriptsize $\pm$1.79} $\downarrow$ & 22.71{\scriptsize $\pm$1.39} $\downarrow$ & 27.69{\scriptsize $\pm$0.96} $\downarrow$ \\
RoTTA {\scriptsize {[CVPR'23]}} & 46.57{\scriptsize $\pm$1.43} $\uparrow$ & 64.95{\scriptsize $\pm$0.39} $\downarrow$ & 51.30{\scriptsize $\pm$0.91} $\uparrow$ & 55.74{\scriptsize $\pm$0.33} $\downarrow$ & 27.25{\scriptsize $\pm$2.14} $\downarrow$ & 44.40{\scriptsize $\pm$0.88} $\uparrow$ & 57.30{\scriptsize $\pm$1.81} $\uparrow$ & 46.10{\scriptsize $\pm$1.70} $\downarrow$ \\
OATTA {\scriptsize {[arXiv'26]}} & 48.39{\scriptsize $\pm$1.50} $\uparrow$ & \underline{69.04}{\scriptsize $\pm$0.64} $\uparrow$ & \underline{53.69}{\scriptsize $\pm$1.48} $\uparrow$ & \underline{57.59}{\scriptsize $\pm$0.25} $\uparrow$ & 27.33{\scriptsize $\pm$3.37} $\downarrow$ & \underline{47.41}{\scriptsize $\pm$0.45} $\uparrow$ & \underline{59.81}{\scriptsize $\pm$3.02} $\uparrow$ & \underline{50.90}{\scriptsize $\pm$3.01} $\uparrow$ \\
TCA {\scriptsize {[ICML'25]}} & 40.92{\scriptsize $\pm$2.26} $\downarrow$ & 41.05{\scriptsize $\pm$0.35} $\downarrow$ & 39.36{\scriptsize $\pm$1.56} $\downarrow$ & 41.07{\scriptsize $\pm$0.27} $\downarrow$ & 20.62{\scriptsize $\pm$0.34} $\downarrow$ & 35.05{\scriptsize $\pm$1.48} $\downarrow$ & 36.13{\scriptsize $\pm$3.79} $\downarrow$ & 36.60{\scriptsize $\pm$1.02} $\downarrow$ \\
REM {\scriptsize {[ICML'25]}} & 46.79{\scriptsize $\pm$1.22} $\uparrow$ & 65.35{\scriptsize $\pm$0.05} $\uparrow$ & 48.77{\scriptsize $\pm$0.26} $\downarrow$ & 54.94{\scriptsize $\pm$0.20} $\downarrow$ & 26.72{\scriptsize $\pm$2.34} $\downarrow$ & 43.91{\scriptsize $\pm$1.46} $\downarrow$ & 58.94{\scriptsize $\pm$3.70} $\uparrow$ & 46.77{\scriptsize $\pm$1.35} $\uparrow$ \\
OFTTA {\scriptsize {[Ubicomp'24]}} & 46.65{\scriptsize $\pm$0.96} $\uparrow$ & 64.78{\scriptsize $\pm$0.73} $\downarrow$ & 48.53{\scriptsize $\pm$2.19} $\downarrow$ & 56.71{\scriptsize $\pm$0.05} $\uparrow$ & 25.79{\scriptsize $\pm$1.52} $\downarrow$ & 44.26{\scriptsize $\pm$0.80} $\downarrow$ & 57.60{\scriptsize $\pm$0.05} $\uparrow$ & 46.35{\scriptsize $\pm$1.76} $\downarrow$ \\
ACCUP {\scriptsize {[KDD'25]}} & 33.15{\scriptsize $\pm$0.35} $\downarrow$ & 33.12{\scriptsize $\pm$0.70} $\downarrow$ & 22.53{\scriptsize $\pm$2.97} $\downarrow$ & 34.12{\scriptsize $\pm$1.37} $\downarrow$ & 29.06{\scriptsize $\pm$0.10} $\uparrow$ & 20.34{\scriptsize $\pm$3.98} $\downarrow$ & 37.58{\scriptsize $\pm$0.78} $\downarrow$ & 21.33{\scriptsize $\pm$1.62} $\downarrow$ \\
COA-HAR {\scriptsize {[ESWA'26]}} & 27.09{\scriptsize $\pm$0.39} $\downarrow$ & 44.77{\scriptsize $\pm$0.31} $\downarrow$ & 27.20{\scriptsize $\pm$1.48} $\downarrow$ & 29.09{\scriptsize $\pm$0.67} $\downarrow$ & 24.66{\scriptsize $\pm$2.31} $\downarrow$ & 20.36{\scriptsize $\pm$0.04} $\downarrow$ & 41.34{\scriptsize $\pm$0.15} $\downarrow$ & 36.08{\scriptsize $\pm$1.37} $\downarrow$ \\
\textbf{\method} (ours) & \textbf{49.74}{\scriptsize $\pm$1.69} $\uparrow$ & \textbf{72.86}{\scriptsize $\pm$0.30} $\uparrow$ & \textbf{56.95}{\scriptsize $\pm$1.62} $\uparrow$ & \textbf{60.23}{\scriptsize $\pm$0.16} $\uparrow$ & \textbf{31.64}{\scriptsize $\pm$0.33} $\uparrow$ & \textbf{50.77}{\scriptsize $\pm$0.48} $\uparrow$ & \textbf{63.30}{\scriptsize $\pm$5.07} $\uparrow$ & \textbf{53.71}{\scriptsize $\pm$3.76} $\uparrow$ \\

\hline
\specialrule{.1em}{0em}{0em}
\end{tabular}
}
\caption{Detailed MF1-scores (\%) of \method and the baselines on additional CAPTURE-24 source--target pairs.}
\label{app:tab:exp:detailed_CAPTURE-24}
\end{table*}

\begin{table*}[p]\small
\centering
\renewcommand{\arraystretch}{1.25}
\setlength{\tabcolsep}{1.7mm}
\resizebox{\linewidth}{!}{%
\begin{tabular}{r||P{1.8cm}P{1.8cm}P{1.8cm}P{1.8cm}P{1.8cm}P{1.8cm}P{1.8cm}P{1.8cm}}
\hline
\specialrule{.1em}{0em}{0em}

\rowcolor{mygray}
Algorithm & S01 $\rightarrow$ S03 & S01 $\rightarrow$ S05 & S02 $\rightarrow$ S01 & S02 $\rightarrow$ S06 & S03 $\rightarrow$ S10 & S03 $\rightarrow$ S13 & S04 $\rightarrow$ S03 & S04 $\rightarrow$ S07 \\
\hline\hline
Source-only & 60.51{\scriptsize $\pm$1.90} $-$ & 45.12{\scriptsize $\pm$8.85} $-$ & 53.25{\scriptsize $\pm$4.60} $-$ & 39.91{\scriptsize $\pm$1.08} $-$ & 40.78{\scriptsize $\pm$1.37} $-$ & 35.74{\scriptsize $\pm$0.72} $-$ & 54.62{\scriptsize $\pm$1.72} $-$ & 52.27{\scriptsize $\pm$3.83} $-$ \\
\hhline{-||-|-|-|-|-|-|-|-}
TENT {\scriptsize {[ICLR'21]}} & 57.52{\scriptsize $\pm$2.55} $\downarrow$ & 43.49{\scriptsize $\pm$6.30} $\downarrow$ & 52.50{\scriptsize $\pm$4.91} $\downarrow$ & \textbf{40.93}{\scriptsize $\pm$0.92} $\uparrow$ & 38.43{\scriptsize $\pm$2.46} $\downarrow$ & 34.96{\scriptsize $\pm$0.92} $\downarrow$ & 53.47{\scriptsize $\pm$2.04} $\downarrow$ & 49.75{\scriptsize $\pm$4.05} $\downarrow$ \\
NOTE {\scriptsize {[NeurIPS'22]}} & 48.52{\scriptsize $\pm$3.63} $\downarrow$ & 46.27{\scriptsize $\pm$4.23} $\uparrow$ & 53.94{\scriptsize $\pm$4.83} $\uparrow$ & 36.61{\scriptsize $\pm$2.56} $\downarrow$ & 35.56{\scriptsize $\pm$2.28} $\downarrow$ & 27.70{\scriptsize $\pm$1.48} $\downarrow$ & 43.51{\scriptsize $\pm$1.72} $\downarrow$ & 43.53{\scriptsize $\pm$2.58} $\downarrow$ \\
SAR {\scriptsize {[ICLR'23]}} & 13.23{\scriptsize $\pm$2.47} $\downarrow$ & 11.55{\scriptsize $\pm$1.14} $\downarrow$ & 14.55{\scriptsize $\pm$3.74} $\downarrow$ & 13.75{\scriptsize $\pm$0.53} $\downarrow$ & 12.23{\scriptsize $\pm$2.10} $\downarrow$ & 13.24{\scriptsize $\pm$1.84} $\downarrow$ & 10.68{\scriptsize $\pm$0.83} $\downarrow$ & 12.54{\scriptsize $\pm$1.21} $\downarrow$ \\
RoTTA {\scriptsize {[CVPR'23]}} & 60.54{\scriptsize $\pm$1.82} $\uparrow$ & 45.10{\scriptsize $\pm$8.88} $\downarrow$ & 53.28{\scriptsize $\pm$4.61} $\uparrow$ & 39.91{\scriptsize $\pm$1.11} $\downarrow$ & 40.70{\scriptsize $\pm$1.41} $\downarrow$ & 35.54{\scriptsize $\pm$0.60} $\downarrow$ & 54.55{\scriptsize $\pm$1.73} $\downarrow$ & 52.41{\scriptsize $\pm$3.90} $\uparrow$ \\
OATTA {\scriptsize {[arXiv'26]}} & \underline{61.03}{\scriptsize $\pm$1.85} $\uparrow$ & \underline{47.51}{\scriptsize $\pm$8.97} $\uparrow$ & \underline{56.14}{\scriptsize $\pm$5.01} $\uparrow$ & \underline{40.90}{\scriptsize $\pm$2.60} $\uparrow$ & \underline{42.46}{\scriptsize $\pm$1.14} $\uparrow$ & \underline{37.44}{\scriptsize $\pm$1.16} $\uparrow$ & \underline{56.08}{\scriptsize $\pm$2.72} $\uparrow$ & \underline{53.70}{\scriptsize $\pm$4.27} $\uparrow$ \\
TCA {\scriptsize {[ICML'25]}} & 28.04{\scriptsize $\pm$6.86} $\downarrow$ & 27.08{\scriptsize $\pm$3.74} $\downarrow$ & 32.73{\scriptsize $\pm$4.46} $\downarrow$ & 20.59{\scriptsize $\pm$5.09} $\downarrow$ & 23.41{\scriptsize $\pm$2.18} $\downarrow$ & 25.18{\scriptsize $\pm$0.43} $\downarrow$ & 32.62{\scriptsize $\pm$1.93} $\downarrow$ & 33.54{\scriptsize $\pm$7.09} $\downarrow$ \\
REM {\scriptsize {[ICML'25]}} & 57.57{\scriptsize $\pm$2.53} $\downarrow$ & 43.53{\scriptsize $\pm$6.04} $\downarrow$ & 52.50{\scriptsize $\pm$4.92} $\downarrow$ & 40.84{\scriptsize $\pm$0.84} $\uparrow$ & 38.13{\scriptsize $\pm$2.52} $\downarrow$ & 34.91{\scriptsize $\pm$0.97} $\downarrow$ & 53.34{\scriptsize $\pm$1.83} $\downarrow$ & 49.66{\scriptsize $\pm$4.10} $\downarrow$ \\
OFTTA {\scriptsize {[Ubicomp'24]}} & 59.75{\scriptsize $\pm$1.52} $\downarrow$ & 45.59{\scriptsize $\pm$9.22} $\uparrow$ & 53.50{\scriptsize $\pm$4.59} $\uparrow$ & 39.93{\scriptsize $\pm$1.10} $\uparrow$ & 40.97{\scriptsize $\pm$1.88} $\uparrow$ & 35.68{\scriptsize $\pm$1.22} $\downarrow$ & 53.90{\scriptsize $\pm$2.31} $\downarrow$ & 52.17{\scriptsize $\pm$3.69} $\downarrow$ \\
ACCUP {\scriptsize {[KDD'25]}} & 27.99{\scriptsize $\pm$3.25} $\downarrow$ & 17.29{\scriptsize $\pm$2.91} $\downarrow$ & 25.52{\scriptsize $\pm$3.33} $\downarrow$ & 24.63{\scriptsize $\pm$4.89} $\downarrow$ & 24.08{\scriptsize $\pm$1.56} $\downarrow$ & 16.09{\scriptsize $\pm$1.49} $\downarrow$ & 20.37{\scriptsize $\pm$2.54} $\downarrow$ & 18.20{\scriptsize $\pm$2.22} $\downarrow$ \\
COA-HAR {\scriptsize {[ESWA'26]}} & 33.26{\scriptsize $\pm$2.57} $\downarrow$ & 36.02{\scriptsize $\pm$4.47} $\downarrow$ & 42.32{\scriptsize $\pm$4.08} $\downarrow$ & 38.02{\scriptsize $\pm$0.43} $\downarrow$ & 31.79{\scriptsize $\pm$1.87} $\downarrow$ & 21.40{\scriptsize $\pm$1.41} $\downarrow$ & 27.01{\scriptsize $\pm$1.28} $\downarrow$ & 27.01{\scriptsize $\pm$6.36} $\downarrow$ \\
\textbf{\method} (ours) & \textbf{62.81}{\scriptsize $\pm$2.03} $\uparrow$ & \textbf{47.84}{\scriptsize $\pm$5.75} $\uparrow$ & \textbf{57.50}{\scriptsize $\pm$5.18} $\uparrow$ & 35.69{\scriptsize $\pm$6.66} $\downarrow$ & \textbf{42.55}{\scriptsize $\pm$1.22} $\uparrow$ & \textbf{37.56}{\scriptsize $\pm$1.28} $\uparrow$ & \textbf{58.87}{\scriptsize $\pm$4.07} $\uparrow$ & \textbf{56.91}{\scriptsize $\pm$6.01} $\uparrow$ \\

\hline
\specialrule{.1em}{0em}{0em}

\rowcolor{mygray}
Algorithm & S04 $\rightarrow$ S09 & S05 $\rightarrow$ S08 & S05 $\rightarrow$ S09 & S05 $\rightarrow$ S12 & S06 $\rightarrow$ S02 & S06 $\rightarrow$ S04 & S07 $\rightarrow$ S05 & S07 $\rightarrow$ S14 \\
\hline\hline
Source-only & 38.81{\scriptsize $\pm$3.77} $-$ & 46.94{\scriptsize $\pm$0.06} $-$ & 55.23{\scriptsize $\pm$4.87} $-$ & 33.00{\scriptsize $\pm$0.01} $-$ & 39.67{\scriptsize $\pm$3.27} $-$ & 39.72{\scriptsize $\pm$3.63} $-$ & 52.61{\scriptsize $\pm$6.31} $-$ & 35.07{\scriptsize $\pm$3.74} $-$ \\
\hhline{-||-|-|-|-|-|-|-|-}
TENT {\scriptsize {[ICLR'21]}} & 36.41{\scriptsize $\pm$4.64} $\downarrow$ & \underline{47.45}{\scriptsize $\pm$0.45} $\uparrow$ & 52.86{\scriptsize $\pm$4.97} $\downarrow$ & 29.68{\scriptsize $\pm$2.56} $\downarrow$ & 39.16{\scriptsize $\pm$2.15} $\downarrow$ & 38.78{\scriptsize $\pm$3.04} $\downarrow$ & 50.78{\scriptsize $\pm$5.46} $\downarrow$ & 33.72{\scriptsize $\pm$3.68} $\downarrow$ \\
NOTE {\scriptsize {[NeurIPS'22]}} & \underline{40.51}{\scriptsize $\pm$5.40} $\uparrow$ & 43.06{\scriptsize $\pm$3.80} $\downarrow$ & 37.33{\scriptsize $\pm$3.49} $\downarrow$ & 30.98{\scriptsize $\pm$2.98} $\downarrow$ & 31.89{\scriptsize $\pm$7.84} $\downarrow$ & 40.28{\scriptsize $\pm$9.81} $\uparrow$ & 46.82{\scriptsize $\pm$4.51} $\downarrow$ & 36.14{\scriptsize $\pm$3.44} $\uparrow$ \\
SAR {\scriptsize {[ICLR'23]}} & 11.02{\scriptsize $\pm$1.14} $\downarrow$ & 9.36{\scriptsize $\pm$0.63} $\downarrow$ & 10.52{\scriptsize $\pm$1.34} $\downarrow$ & 9.12{\scriptsize $\pm$2.43} $\downarrow$ & 12.43{\scriptsize $\pm$2.08} $\downarrow$ & 8.86{\scriptsize $\pm$0.27} $\downarrow$ & 11.22{\scriptsize $\pm$0.71} $\downarrow$ & 10.41{\scriptsize $\pm$0.80} $\downarrow$ \\
RoTTA {\scriptsize {[CVPR'23]}} & 38.68{\scriptsize $\pm$3.81} $\downarrow$ & 46.81{\scriptsize $\pm$0.25} $\downarrow$ & 55.32{\scriptsize $\pm$5.00} $\uparrow$ & 32.86{\scriptsize $\pm$0.42} $\downarrow$ & 39.83{\scriptsize $\pm$3.21} $\uparrow$ & 39.92{\scriptsize $\pm$3.80} $\uparrow$ & 52.00{\scriptsize $\pm$5.91} $\downarrow$ & 34.77{\scriptsize $\pm$3.57} $\downarrow$ \\
OATTA {\scriptsize {[arXiv'26]}} & 40.30{\scriptsize $\pm$4.02} $\uparrow$ & \textbf{48.99}{\scriptsize $\pm$0.25} $\uparrow$ & \underline{57.94}{\scriptsize $\pm$5.13} $\uparrow$ & \textbf{34.05}{\scriptsize $\pm$0.41} $\uparrow$ & \underline{40.70}{\scriptsize $\pm$3.97} $\uparrow$ & \underline{40.45}{\scriptsize $\pm$3.88} $\uparrow$ & 54.32{\scriptsize $\pm$5.80} $\uparrow$ & \underline{37.26}{\scriptsize $\pm$3.92} $\uparrow$ \\
TCA {\scriptsize {[ICML'25]}} & 23.82{\scriptsize $\pm$2.71} $\downarrow$ & 21.77{\scriptsize $\pm$1.61} $\downarrow$ & 31.77{\scriptsize $\pm$4.69} $\downarrow$ & 20.34{\scriptsize $\pm$1.31} $\downarrow$ & 27.77{\scriptsize $\pm$2.07} $\downarrow$ & 31.46{\scriptsize $\pm$0.95} $\downarrow$ & 30.40{\scriptsize $\pm$4.89} $\downarrow$ & 15.71{\scriptsize $\pm$5.09} $\downarrow$ \\
REM {\scriptsize {[ICML'25]}} & 36.44{\scriptsize $\pm$4.61} $\downarrow$ & 47.30{\scriptsize $\pm$0.42} $\uparrow$ & 52.59{\scriptsize $\pm$4.77} $\downarrow$ & 29.59{\scriptsize $\pm$2.57} $\downarrow$ & 38.91{\scriptsize $\pm$2.20} $\downarrow$ & 38.75{\scriptsize $\pm$2.80} $\downarrow$ & 50.93{\scriptsize $\pm$5.43} $\downarrow$ & 33.48{\scriptsize $\pm$3.65} $\downarrow$ \\
OFTTA {\scriptsize {[Ubicomp'24]}} & 39.15{\scriptsize $\pm$4.20} $\uparrow$ & 46.93{\scriptsize $\pm$0.10} $\downarrow$ & 55.14{\scriptsize $\pm$4.90} $\downarrow$ & \underline{33.25}{\scriptsize $\pm$0.93} $\uparrow$ & 39.66{\scriptsize $\pm$3.37} $\downarrow$ & 39.91{\scriptsize $\pm$4.26} $\uparrow$ & \underline{54.66}{\scriptsize $\pm$7.96} $\uparrow$ & 35.11{\scriptsize $\pm$3.71} $\uparrow$ \\
ACCUP {\scriptsize {[KDD'25]}} & 22.59{\scriptsize $\pm$3.90} $\downarrow$ & 18.78{\scriptsize $\pm$0.51} $\downarrow$ & 25.78{\scriptsize $\pm$3.26} $\downarrow$ & 15.73{\scriptsize $\pm$5.05} $\downarrow$ & 23.79{\scriptsize $\pm$1.28} $\downarrow$ & 24.82{\scriptsize $\pm$2.17} $\downarrow$ & 21.05{\scriptsize $\pm$3.68} $\downarrow$ & 14.54{\scriptsize $\pm$2.48} $\downarrow$ \\
COA-HAR {\scriptsize {[ESWA'26]}} & 32.45{\scriptsize $\pm$3.42} $\downarrow$ & 30.25{\scriptsize $\pm$2.26} $\downarrow$ & 35.65{\scriptsize $\pm$4.99} $\downarrow$ & 25.84{\scriptsize $\pm$3.75} $\downarrow$ & 31.30{\scriptsize $\pm$12.58} $\downarrow$ & 34.26{\scriptsize $\pm$11.19} $\downarrow$ & 30.59{\scriptsize $\pm$4.41} $\downarrow$ & 20.97{\scriptsize $\pm$4.45} $\downarrow$ \\
\textbf{\method} (ours) & \textbf{40.73}{\scriptsize $\pm$3.02} $\uparrow$ & 44.05{\scriptsize $\pm$5.77} $\downarrow$ & \textbf{58.11}{\scriptsize $\pm$5.54} $\uparrow$ & 31.87{\scriptsize $\pm$3.07} $\downarrow$ & \textbf{40.96}{\scriptsize $\pm$3.00} $\uparrow$ & \textbf{40.89}{\scriptsize $\pm$2.35} $\uparrow$ & \textbf{55.74}{\scriptsize $\pm$3.14} $\uparrow$ & \textbf{39.19}{\scriptsize $\pm$4.21} $\uparrow$ \\

\hline
\specialrule{.1em}{0em}{0em}
\end{tabular}
}
\caption{Detailed MF1-scores (\%) of \method and the baselines on additional USC-HAD source--target pairs.}
\label{app:tab:exp:detailed_USC-HAD}
\end{table*}

\begin{table*}[p]\small
\centering
\renewcommand{\arraystretch}{1.25}
\setlength{\tabcolsep}{1.7mm}
\resizebox{\linewidth}{!}{%
\begin{tabular}{r||P{1.8cm}P{1.8cm}P{1.8cm}P{1.8cm}P{1.8cm}P{1.8cm}P{1.8cm}P{1.8cm}}
\hline
\specialrule{.1em}{0em}{0em}

\rowcolor{mygray}
Algorithm & S01 $\rightarrow$ S16 & S01 $\rightarrow$ S26 & S02 $\rightarrow$ S22 & S02 $\rightarrow$ S24 & S03 $\rightarrow$ S25 & S03 $\rightarrow$ S28 & S04 $\rightarrow$ S17 & S04 $\rightarrow$ S20 \\
\hline\hline
Source-only & 35.26{\scriptsize $\pm$3.50} $-$ & 32.38{\scriptsize $\pm$6.27} $-$ & 70.78{\scriptsize $\pm$4.93} $-$ & 79.42{\scriptsize $\pm$0.82} $-$ & 61.86{\scriptsize $\pm$5.71} $-$ & 59.90{\scriptsize $\pm$3.24} $-$ & 59.54{\scriptsize $\pm$3.98} $-$ & 43.36{\scriptsize $\pm$7.33} $-$ \\
\hhline{-||-|-|-|-|-|-|-|-}
TENT {\scriptsize {[ICLR'21]}} & 35.19{\scriptsize $\pm$1.70} $\downarrow$ & 31.55{\scriptsize $\pm$4.62} $\downarrow$ & 73.62{\scriptsize $\pm$1.42} $\uparrow$ & 77.16{\scriptsize $\pm$0.60} $\downarrow$ & 58.90{\scriptsize $\pm$6.69} $\downarrow$ & 60.89{\scriptsize $\pm$4.46} $\uparrow$ & 56.49{\scriptsize $\pm$2.17} $\downarrow$ & 49.26{\scriptsize $\pm$12.78} $\uparrow$ \\
NOTE {\scriptsize {[NeurIPS'22]}} & 34.96{\scriptsize $\pm$4.51} $\downarrow$ & 41.82{\scriptsize $\pm$7.41} $\uparrow$ & 71.91{\scriptsize $\pm$7.20} $\uparrow$ & \underline{82.81}{\scriptsize $\pm$1.80} $\uparrow$ & 64.92{\scriptsize $\pm$3.41} $\uparrow$ & \underline{63.05}{\scriptsize $\pm$4.37} $\uparrow$ & \textbf{60.36}{\scriptsize $\pm$4.46} $\uparrow$ & 43.74{\scriptsize $\pm$7.15} $\uparrow$ \\
SAR {\scriptsize {[ICLR'23]}} & 21.03{\scriptsize $\pm$0.88} $\downarrow$ & 21.44{\scriptsize $\pm$0.99} $\downarrow$ & 26.61{\scriptsize $\pm$3.26} $\downarrow$ & 23.48{\scriptsize $\pm$5.53} $\downarrow$ & 11.34{\scriptsize $\pm$2.76} $\downarrow$ & 16.56{\scriptsize $\pm$1.88} $\downarrow$ & 15.84{\scriptsize $\pm$5.16} $\downarrow$ & 19.65{\scriptsize $\pm$1.89} $\downarrow$ \\
RoTTA {\scriptsize {[CVPR'23]}} & 35.26{\scriptsize $\pm$3.50} $\downarrow$ & 32.38{\scriptsize $\pm$6.27} $\downarrow$ & 70.71{\scriptsize $\pm$4.88} $\downarrow$ & 79.36{\scriptsize $\pm$0.90} $\downarrow$ & 61.86{\scriptsize $\pm$5.71} $\downarrow$ & 59.90{\scriptsize $\pm$3.24} $\downarrow$ & 59.54{\scriptsize $\pm$3.98} $\downarrow$ & 43.36{\scriptsize $\pm$7.33} $\downarrow$ \\
OATTA {\scriptsize {[arXiv'26]}} & 33.48{\scriptsize $\pm$3.98} $\downarrow$ & 32.57{\scriptsize $\pm$3.44} $\uparrow$ & 71.76{\scriptsize $\pm$6.16} $\uparrow$ & 79.85{\scriptsize $\pm$2.08} $\uparrow$ & \underline{69.23}{\scriptsize $\pm$6.70} $\uparrow$ & 61.79{\scriptsize $\pm$1.95} $\uparrow$ & 56.50{\scriptsize $\pm$4.05} $\downarrow$ & 44.51{\scriptsize $\pm$7.47} $\uparrow$ \\
TCA {\scriptsize {[ICML'25]}} & 21.42{\scriptsize $\pm$0.79} $\downarrow$ & 28.70{\scriptsize $\pm$8.10} $\downarrow$ & 57.15{\scriptsize $\pm$12.27} $\downarrow$ & 54.99{\scriptsize $\pm$4.67} $\downarrow$ & 33.77{\scriptsize $\pm$5.32} $\downarrow$ & 38.18{\scriptsize $\pm$6.55} $\downarrow$ & 50.76{\scriptsize $\pm$2.30} $\downarrow$ & 28.76{\scriptsize $\pm$7.34} $\downarrow$ \\
REM {\scriptsize {[ICML'25]}} & 34.52{\scriptsize $\pm$1.04} $\downarrow$ & 31.22{\scriptsize $\pm$4.57} $\downarrow$ & \textbf{74.80}{\scriptsize $\pm$0.82} $\uparrow$ & 77.86{\scriptsize $\pm$0.50} $\downarrow$ & 58.90{\scriptsize $\pm$6.29} $\downarrow$ & 60.40{\scriptsize $\pm$5.00} $\uparrow$ & 56.35{\scriptsize $\pm$1.89} $\downarrow$ & 48.38{\scriptsize $\pm$11.49} $\uparrow$ \\
OFTTA {\scriptsize {[Ubicomp'24]}} & \underline{36.84}{\scriptsize $\pm$3.61} $\uparrow$ & 33.62{\scriptsize $\pm$7.72} $\uparrow$ & 68.00{\scriptsize $\pm$4.76} $\downarrow$ & 79.51{\scriptsize $\pm$1.40} $\uparrow$ & 58.15{\scriptsize $\pm$4.92} $\downarrow$ & 59.13{\scriptsize $\pm$4.89} $\downarrow$ & \underline{59.67}{\scriptsize $\pm$4.11} $\uparrow$ & 44.68{\scriptsize $\pm$8.44} $\uparrow$ \\
ACCUP {\scriptsize {[KDD'25]}} & 34.47{\scriptsize $\pm$2.32} $\downarrow$ & \textbf{49.63}{\scriptsize $\pm$3.40} $\uparrow$ & 71.29{\scriptsize $\pm$3.16} $\uparrow$ & 69.71{\scriptsize $\pm$5.35} $\downarrow$ & 55.22{\scriptsize $\pm$2.58} $\downarrow$ & 52.08{\scriptsize $\pm$1.98} $\downarrow$ & 48.15{\scriptsize $\pm$3.27} $\downarrow$ & \underline{49.41}{\scriptsize $\pm$5.55} $\uparrow$ \\
COA-HAR {\scriptsize {[ESWA'26]}} & 35.27{\scriptsize $\pm$0.02} $\uparrow$ & \underline{43.35}{\scriptsize $\pm$3.95} $\uparrow$ & 59.03{\scriptsize $\pm$6.62} $\downarrow$ & 66.64{\scriptsize $\pm$6.10} $\downarrow$ & 36.35{\scriptsize $\pm$3.19} $\downarrow$ & 47.04{\scriptsize $\pm$2.03} $\downarrow$ & 50.55{\scriptsize $\pm$1.66} $\downarrow$ & 40.56{\scriptsize $\pm$9.39} $\downarrow$ \\
\textbf{\method} (ours) & \textbf{39.34}{\scriptsize $\pm$2.74} $\uparrow$ & 42.86{\scriptsize $\pm$1.30} $\uparrow$ & \underline{73.81}{\scriptsize $\pm$4.56} $\uparrow$ & \textbf{83.86}{\scriptsize $\pm$2.54} $\uparrow$ & \textbf{72.19}{\scriptsize $\pm$8.01} $\uparrow$ & \textbf{70.35}{\scriptsize $\pm$1.75} $\uparrow$ & 57.34{\scriptsize $\pm$2.35} $\downarrow$ & \textbf{52.45}{\scriptsize $\pm$4.33} $\uparrow$ \\

\hline
\specialrule{.1em}{0em}{0em}

\rowcolor{mygray}
Algorithm & S05 $\rightarrow$ S18 & S05 $\rightarrow$ S29 & S07 $\rightarrow$ S19 & S07 $\rightarrow$ S27 & S08 $\rightarrow$ S21 & S08 $\rightarrow$ S23 & S09 $\rightarrow$ S19 & S09 $\rightarrow$ S30 \\
\hline\hline
Source-only & 54.06{\scriptsize $\pm$3.62} $-$ & 71.01{\scriptsize $\pm$5.32} $-$ & 50.11{\scriptsize $\pm$3.12} $-$ & 73.79{\scriptsize $\pm$0.59} $-$ & 78.67{\scriptsize $\pm$2.82} $-$ & 75.29{\scriptsize $\pm$4.81} $-$ & 40.33{\scriptsize $\pm$4.06} $-$ & 34.63{\scriptsize $\pm$2.63} $-$ \\
\hhline{-||-|-|-|-|-|-|-|-}
TENT {\scriptsize {[ICLR'21]}} & 52.58{\scriptsize $\pm$4.62} $\downarrow$ & 68.44{\scriptsize $\pm$2.31} $\downarrow$ & 51.07{\scriptsize $\pm$3.92} $\uparrow$ & 71.40{\scriptsize $\pm$1.43} $\downarrow$ & 74.64{\scriptsize $\pm$2.88} $\downarrow$ & 73.46{\scriptsize $\pm$3.49} $\downarrow$ & \underline{41.59}{\scriptsize $\pm$5.38} $\uparrow$ & 37.33{\scriptsize $\pm$3.98} $\uparrow$ \\
NOTE {\scriptsize {[NeurIPS'22]}} & \underline{58.86}{\scriptsize $\pm$2.60} $\uparrow$ & 72.34{\scriptsize $\pm$3.35} $\uparrow$ & \underline{52.55}{\scriptsize $\pm$2.58} $\uparrow$ & 75.00{\scriptsize $\pm$1.94} $\uparrow$ & \textbf{80.46}{\scriptsize $\pm$1.07} $\uparrow$ & \underline{76.72}{\scriptsize $\pm$4.88} $\uparrow$ & 40.57{\scriptsize $\pm$1.51} $\uparrow$ & \underline{38.93}{\scriptsize $\pm$2.09} $\uparrow$ \\
SAR {\scriptsize {[ICLR'23]}} & 14.87{\scriptsize $\pm$0.15} $\downarrow$ & 17.01{\scriptsize $\pm$2.29} $\downarrow$ & 23.58{\scriptsize $\pm$1.77} $\downarrow$ & 30.25{\scriptsize $\pm$3.39} $\downarrow$ & 31.75{\scriptsize $\pm$0.15} $\downarrow$ & 22.31{\scriptsize $\pm$1.25} $\downarrow$ & 25.49{\scriptsize $\pm$2.02} $\downarrow$ & 17.70{\scriptsize $\pm$1.99} $\downarrow$ \\
RoTTA {\scriptsize {[CVPR'23]}} & 54.06{\scriptsize $\pm$3.62} $\downarrow$ & 71.01{\scriptsize $\pm$5.32} $\downarrow$ & 50.11{\scriptsize $\pm$3.12} $\downarrow$ & 73.71{\scriptsize $\pm$0.57} $\downarrow$ & 78.67{\scriptsize $\pm$2.82} $\downarrow$ & 75.29{\scriptsize $\pm$4.81} $\downarrow$ & 40.33{\scriptsize $\pm$4.06} $\downarrow$ & 34.55{\scriptsize $\pm$2.65} $\downarrow$ \\
OATTA {\scriptsize {[arXiv'26]}} & 53.38{\scriptsize $\pm$5.99} $\downarrow$ & \underline{72.87}{\scriptsize $\pm$6.71} $\uparrow$ & 48.58{\scriptsize $\pm$2.24} $\downarrow$ & \underline{76.51}{\scriptsize $\pm$1.32} $\uparrow$ & 78.78{\scriptsize $\pm$4.21} $\uparrow$ & 76.29{\scriptsize $\pm$5.27} $\uparrow$ & 37.29{\scriptsize $\pm$5.92} $\downarrow$ & 31.68{\scriptsize $\pm$2.85} $\downarrow$ \\
TCA {\scriptsize {[ICML'25]}} & 23.47{\scriptsize $\pm$7.20} $\downarrow$ & 54.36{\scriptsize $\pm$4.49} $\downarrow$ & 46.76{\scriptsize $\pm$8.96} $\downarrow$ & 61.30{\scriptsize $\pm$6.25} $\downarrow$ & 61.65{\scriptsize $\pm$6.07} $\downarrow$ & 48.29{\scriptsize $\pm$12.11} $\downarrow$ & 34.39{\scriptsize $\pm$0.73} $\downarrow$ & 18.03{\scriptsize $\pm$5.29} $\downarrow$ \\
REM {\scriptsize {[ICML'25]}} & 51.42{\scriptsize $\pm$4.02} $\downarrow$ & 68.76{\scriptsize $\pm$2.30} $\downarrow$ & 51.21{\scriptsize $\pm$4.58} $\uparrow$ & 70.92{\scriptsize $\pm$1.00} $\downarrow$ & 74.47{\scriptsize $\pm$2.82} $\downarrow$ & 72.34{\scriptsize $\pm$3.35} $\downarrow$ & 41.44{\scriptsize $\pm$5.14} $\uparrow$ & 37.03{\scriptsize $\pm$4.07} $\uparrow$ \\
OFTTA {\scriptsize {[Ubicomp'24]}} & 53.54{\scriptsize $\pm$3.13} $\downarrow$ & 67.04{\scriptsize $\pm$4.50} $\downarrow$ & 49.96{\scriptsize $\pm$3.17} $\downarrow$ & 70.91{\scriptsize $\pm$0.88} $\downarrow$ & \underline{79.35}{\scriptsize $\pm$1.52} $\uparrow$ & 75.38{\scriptsize $\pm$5.53} $\uparrow$ & 40.79{\scriptsize $\pm$3.31} $\uparrow$ & 36.78{\scriptsize $\pm$4.51} $\uparrow$ \\
ACCUP {\scriptsize {[KDD'25]}} & 52.98{\scriptsize $\pm$0.87} $\downarrow$ & 64.49{\scriptsize $\pm$4.65} $\downarrow$ & 44.89{\scriptsize $\pm$7.76} $\downarrow$ & 67.70{\scriptsize $\pm$2.06} $\downarrow$ & 63.41{\scriptsize $\pm$1.21} $\downarrow$ & 73.78{\scriptsize $\pm$5.56} $\downarrow$ & 38.89{\scriptsize $\pm$3.63} $\downarrow$ & 36.31{\scriptsize $\pm$4.70} $\uparrow$ \\
COA-HAR {\scriptsize {[ESWA'26]}} & 31.49{\scriptsize $\pm$3.82} $\downarrow$ & 58.38{\scriptsize $\pm$5.29} $\downarrow$ & 45.18{\scriptsize $\pm$4.91} $\downarrow$ & 56.23{\scriptsize $\pm$2.37} $\downarrow$ & 71.83{\scriptsize $\pm$3.19} $\downarrow$ & 64.70{\scriptsize $\pm$8.88} $\downarrow$ & 40.62{\scriptsize $\pm$4.08} $\uparrow$ & 28.83{\scriptsize $\pm$3.44} $\downarrow$ \\
\textbf{\method} (ours) & \textbf{59.89}{\scriptsize $\pm$4.33} $\uparrow$ & \textbf{73.04}{\scriptsize $\pm$4.53} $\uparrow$ & \textbf{53.22}{\scriptsize $\pm$3.04} $\uparrow$ & \textbf{80.05}{\scriptsize $\pm$1.52} $\uparrow$ & 76.84{\scriptsize $\pm$4.29} $\downarrow$ & \textbf{77.89}{\scriptsize $\pm$3.18} $\uparrow$ & \textbf{46.20}{\scriptsize $\pm$5.10} $\uparrow$ & \textbf{40.03}{\scriptsize $\pm$5.72} $\uparrow$ \\

\hline
\specialrule{.1em}{0em}{0em}
\end{tabular}
}
\caption{Detailed MF1-scores (\%) of \method and the baselines on additional UCI-HAR source--target pairs.}
\label{app:tab:exp:detailed_UCI-HAR}
\end{table*}

\end{document}